\documentclass[letterpaper]{article} % DO NOT CHANGE THIS
\newcommand{\versionnote}{Work in progress. This version: November 9, 2026} 
\usepackage[preprint]{aaai2027}  % arXiv preprint: authors shown, no AAAI copyright slug
\usepackage[hyphens]{url}  % DO NOT CHANGE THIS
\usepackage{graphicx} % DO NOT CHANGE THIS
\usepackage{natbib}  % DO NOT CHANGE THIS AND DO NOT ADD ANY OPTIONS TO IT
\usepackage{caption} % DO NOT CHANGE THIS AND DO NOT ADD ANY OPTIONS TO IT
\usepackage{booktabs}
\usepackage{longtable} % Task inventory in the appendix
\usepackage{amsmath}
\usepackage{amssymb}
\usepackage{xcolor}
\usepackage{tikz}
\usepackage{fontawesome5}
\usetikzlibrary{positioning, fit, calc, arrows.meta, backgrounds}

\newcommand{\bench}{\textsc{Ask2Decide}}
\newcommand{\op}{\textsc{CalOp}}          % human-calibrated simulated operator (code name f_min4)
\newcommand{\defop}{\textsc{DefaultOp}}   % one-prompt LLM operator (code name default)

\definecolor{humancol}{rgb}{0.18,0.31,0.31}
\definecolor{agentcol}{RGB}{200,90,20}
\definecolor{solvecol}{RGB}{70,90,160}
\definecolor{goodcol}{RGB}{40,140,70}
\definecolor{badcol}{RGB}{190,40,40}
\tikzset{
  box/.style={draw=humancol, rounded corners=2pt, fill=white, align=center,
              font=\scriptsize, inner sep=3pt, text width=#1},
  box/.default={3.6cm},
  soft/.style={box=#1, fill=humancol!10},
  soft/.default={3.6cm},
  agent/.style={box=#1, draw=agentcol, fill=agentcol!10},
  agent/.default={3.6cm},
  grp/.style={draw=humancol, dashed, rounded corners=4pt, inner sep=5pt},
  flow/.style={-{Stealth[length=2.2mm]}, thick, humancol},
  lbl/.style={font=\scriptsize, align=center, text=black},
  tag/.style={font=\scriptsize\bfseries, text=humancol},
  solver/.style={box=#1, draw=solvecol, fill=solvecol!10},
  solver/.default={3.6cm},
  aflow/.style={-{Stealth[length=2.2mm]}, thick, agentcol},
  rflow/.style={-{Stealth[length=2.2mm]}, thick, badcol},
  ptitle/.style={font=\footnotesize\bfseries, text=#1!80!black},
  qbub/.style={draw=agentcol, fill=agentcol!10, rounded corners=5pt, font=\scriptsize, inner sep=2.5pt, align=left},
  abub/.style={draw=humancol, fill=humancol!10, rounded corners=5pt, font=\scriptsize, inner sep=2.5pt, align=left},
  note/.style={draw=humancol, fill=yellow!12, font=\scriptsize, inner sep=2.5pt, align=left, text width=#1},
  note/.default={2.6cm},
}

\usepackage{eso-pic}
\newcommand{\versionstamp}{%
  \AddToShipoutPictureFG*{\AtPageUpperLeft{%
    \put(\LenToUnit{\paperwidth-0.75in},\LenToUnit{-0.5in}){%
      \makebox[0pt][r]{\footnotesize\itshape \versionnote}}}}}

\newcounter{seniorfn}
\newcommand{\seniorauthor}{%
  \ifnum\value{seniorfn}=0%
    \footnote{Senior authors.}\setcounter{seniorfn}{\value{footnote}}%
  \else%
    \footnotemark[\value{seniorfn}]%
  \fi}

\title{From Questions to Decisions: Benchmarking How LLM Agents \\ Acquire Decision-Relevant Information from Human Beings}

\author{
    Yuhan Tang\textsuperscript{\rm 1}\equalcontrib,
    Ke Fan\textsuperscript{\rm 2}\equalcontrib,
    Xinkai Zou\textsuperscript{\rm 3}\equalcontrib,
    Baichuan Mo\textsuperscript{\rm 2}\corresponding,
    Kangxin Cui\textsuperscript{\rm 4},
        Ao Qu\textsuperscript{\rm 1}, \\
        Ou Zheng\textsuperscript{\rm 5},
    Jiangbo Yu\textsuperscript{\rm 6}, Xuan Jiang\textsuperscript{\rm 1},
    Haris N. Koutsopoulos\textsuperscript{\rm 7},
    Jinhua Zhao\textsuperscript{\rm 1}\seniorauthor,
    Yuan Zhou\textsuperscript{\rm 2}\seniorauthor
}
\affiliations{
    \textsuperscript{\rm 1}Massachusetts Institute of Technology \quad
    \textsuperscript{\rm 2}Tsinghua University \quad
    \textsuperscript{\rm 3}University of California, San Diego\\
    \textsuperscript{\rm 4}Tongji University\quad
    \textsuperscript{\rm
    5}Peking University\quad
    \textsuperscript{\rm
    6}McGill University\quad
    \textsuperscript{\rm 7}Northeastern University
}

\makeatletter
\let\arxiv@originalPackageError\PackageError
\long\def\PackageError#1#2#3{%
  \begingroup
  \def\arxiv@package{#1}%
  \def\arxiv@aaai{aaai}%
  \def\arxiv@message{#2}%
  \def\arxiv@hyperrefban{You must not use hyperref in AAAI papers.}%
  \ifx\arxiv@package\arxiv@aaai
    \ifx\arxiv@message\arxiv@hyperrefban
      \PackageInfo{arxiv_version}{Hyperlinks enabled for the arXiv preprint}%
    \else
      \arxiv@originalPackageError{#1}{#2}{#3}%
    \fi
  \else
    \arxiv@originalPackageError{#1}{#2}{#3}%
  \fi
  \endgroup}
\AtBeginDocument{\let\PackageError\arxiv@originalPackageError}
\makeatother
\usepackage[hidelinks,bookmarksnumbered=true]{hyperref}
\begin{document}
\versionstamp
\maketitle

% ===========================================================================
\begin{abstract}
% Effective planning requires agents to obtain knowledge that people hold but do not state in the initial request.
% Existing interactive benchmarks often represent this knowledge through simulated users.
% Yet these users can disclose information more readily than real people, making tasks easier and overstating agent performance.
% Reliable evaluation therefore depends on how closely simulated interactions reflect human behavior.
% We introduce \bench{}, a benchmark and human-alignment pipeline for evaluating planning agents that must ask people for missing information.
% The benchmark measures whether an agent obtains the information needed to produce a feasible, high-quality plan.
% The pipeline uses real interaction data to calibrate simulated operators and evaluates both their responses and the resulting planning outcomes.
% Using 79 sessions from 19 participants, we develop an operator that more closely matches observed human response patterns than a default LLM simulator.
% The benchmark contains 43 tasks. On a 24-task subset, initial-plan infeasibility is 58--67\% when the agent must ask for information, compared with 4--8\% when all information is provided upfront.
% These results expose a gap between solving a fully specified problem and obtaining the information needed to solve it.
% Together, the benchmark and alignment pipeline support more realistic evaluation of agents that depend on human knowledge.

LLM agents are increasingly asked to make plans for people, such as scheduling drivers or assigning nurses.
However, people's requests are often incomplete, and the supporting information for decision making may be outdated, requiring agents to ask for clarification.
Existing benchmarks typically provide fully specified problems or rely on uncalibrated LLM-simulated users, leaving this ability insufficiently tested.
We introduce \bench{}, a benchmark of 43 planning tasks that evaluates whether agents ask for missing information that matters for their decisions.
An operator holds facts absent from the agent's request and records, and optimization solvers determine which facts change the optimal plan.
We also include similarly worded facts that leave the optimal plan unchanged to test whether agents distinguish useful information from irrelevant details.
We collect 79 interaction sessions from 19 participants and find that, after an incorrect plan, human replies omit the correcting fact far more often than replies from an uncalibrated simulated user (43\% versus 7\%).
Using these interactions, we calibrate a simulated operator that more closely matches human reply patterns.
Evaluating four frontier LLMs, we find that their initial plans are infeasible on 58--98\% of tasks when information is missing, compared with 5--28\% when all facts are provided upfront.
Agents uncover facts that change the optimal plan only slightly more often than facts that leave it unchanged (39--52\% versus 32--41\%).
These results show that obtaining decision-relevant information through interaction remains a major bottleneck for LLM planning.
\textbf{This is ongoing work and will be updated. The date of this version is shown on the first page.}
\end{abstract}

% ===========================================================================
\section{Introduction}\label{sec:intro}

LLM agents are increasingly asked to make plans for real-world operations. Recent work has made progress in
translating complete problem descriptions into optimization
models~\citep{ramamonjison2023nl4opt,ahmaditeshnizi2024optimus,huang2025orlm,jiang2025llmopt}. In practice, however, neither the initial request nor the organization’s records describe the problem as it stands today. Some rules were never written down; some conditions changed after the records were made, through an accident, new information or a system update not yet entered; and some records are simply wrong. Those missing information could make the final plans infeasible. A dispatcher asking for today's delivery routes may
omit that one driver must return by 10:00. Assigning that driver a longer route could make the plan infeasible.
Such conditions are known to the people involved but absent from the request and the available records, so the
agent can learn them only from those people. Yet missing facts differ in their consequences: some change which plan is feasible or
preferred, while others leave the decision unchanged. Effective planning therefore requires the agent to recognize
what it needs to know and direct its questions toward those facts. This raises a question that current benchmarks
do not answer: can an agent elicit the missing information that matters for its decision?

Existing evaluations fall short in two aspects.
(1)~\emph{They do not measure the decision value of acquired information.}
Existing optimization evaluations focus on modeling and solving from fixed problem descriptions, without testing
whether agents uncover temporary changes in operational conditions. Benchmarks for clarification and information
acquisition address missing information: QuestBench tests whether models identify the clarification question needed
to solve an underspecified task~\citep{li2025questbench}, while OR-Clarify evaluates the recovery of withheld
formulation requirements before modeling~\citep{orclarify2026}. However, these evaluations do not quantify the
consequences of missing facts for plan feasibility and solution quality, or evaluate the operational plan produced
after the interaction. This leaves open whether agents acquire the information that matters for the decisions
they make.
(2)~\emph{Their simulated users do not reliably communicate as real human beings do.}
Interactive benchmarks evaluate agents through conversations with LLM-simulated users~\citep{yao2024taubench}.
Compared with real human beings, default simulators produce overly verbose and polite replies, exhibit less
stylistic variation, and disclose information more completely and earlier in the
conversation~\citep{dou2025simulatorarena,zhou2026sim2real}. These differences can change how much agents need to ask
and yield success rates and failure patterns that differ from human evaluations~\citep{seshadri2026lost}.
Reliable evaluation therefore requires calibrating and validating simulated responses against real human
interactions.

To address both gaps, we introduce \bench{}, a benchmark for evaluating how agents acquire decision-relevant
information from people. It contains 43 planning tasks in routing, scheduling, staffing and other domains.
The agent receives a brief and an incomplete knowledge base, and an operator holds today's hidden facts.
Optimization makes a good testbed because a solver can do two things exactly. It computes the value of each hidden
fact: {its {regret} $\Delta$ is the loss that the safest plan can still incur when the agent does not know whether that fact holds, with the rest of the problem fixed.} And it scores
any submitted plan against the true facts, from 0 for an optimal plan to 1 for a plan that cannot be carried out.
Facts that change the plan ($\Delta>0$) are paired with \emph{twins}: facts of the same form about another object
that sound just as important but leave the best plan unchanged. Figure~\ref{fig:example}a shows such a pair. Recovering many facts
is therefore not the same as asking useful questions.

Table~\ref{tab:motivation-intro} runs the same planning agent (DeepSeek V4.1 Flash) on 22 tasks, once with the human operators of our
study (19 participants, 79 sessions) and once with a default LLM operator (three runs per task). First, interactive planning is hard:
with a full brief (every hidden fact in the first message), the initial plan is rarely infeasible (3.0--6.3\%), but in
the interactive condition, where the hidden facts are not given, it is infeasible in 32.9--48.5\% of episodes. Rejections
and revisions bring the final plan below 8\%. Second, the operator changes
the result: the initial plan fails more often with the default operator than with people, but the final plan fails
less often, because after a wrong plan the default operator withholds the correcting fact in only 7\% of replies,
against 43\% for people. We therefore build \op{}, a simulated operator calibrated on the human sessions, and
Section~\ref{sec:human} validates it against them.

\begin{table}[t]
\centering
\small
\setlength{\tabcolsep}{3pt}
\begin{tabular}{lcccc}
\toprule
 & \multicolumn{2}{c}{Default LLM operator} & \multicolumn{2}{c}{Human operators} \\
\cmidrule(lr){2-3}\cmidrule(lr){4-5}
 & Initial & Final & Initial & Final \\
\midrule
Interactive & 48.5\% & 6.1\% & 32.9\% & 7.6\% \\
Full brief  & 3.0\%  & 0.0\% & 6.3\% & 3.8\% \\
\bottomrule
\end{tabular}
\caption{Share of infeasible plans with a default LLM operator and with human operators (22 tasks, planning agent:
DeepSeek V4.1 Flash).}
\label{tab:motivation-intro}
\end{table}

Because a solver can score every plan both with and without the hidden facts, \bench{} can separate abilities
that a single success rate mixes together: modelling the problem, asking for the facts that matter, and
recovering after a rejected plan. Since the operator is simulated, we also test how the choice of operator changes
these measurements. In Section~\ref{sec:agents} we use \op{} to evaluate four LLMs as the planning agent on all 43 tasks and
organize the results around four research questions.

\emph{RQ1: How does access to complete information affect initial-plan quality?}
Strongly. Models that plan well when told every hidden fact produce an initial plan that cannot be carried out on
most tasks when those facts are not given. Obtaining the information, more than modelling the problem, is the bottleneck.

\emph{RQ2: Do models ask about facts that change the plan?}
Only weakly. Models uncover the facts that change the plan barely more often than look-alike twins that do not,
and on large instances asking more questions does not rescue the initial plan.

\emph{RQ3: How much do models recover after a rejected plan?}
Much, but recovery hides how they asked. A rejection itself tells the agent that something is wrong today, and
after revisions, models that asked very differently end up close together. We therefore score the initial plan for
asking and report the final plan as a measure of recovery.

\emph{RQ4: How does the choice of simulated operator affect planning outcomes?}
It shifts individual scores in both directions but not the main conclusion: under both a default and the
calibrated operator, the interactive condition leaves far more initial plans infeasible than a full brief.

In summary, our contributions are threefold:
\begin{itemize}
  \item We introduce \bench{}, a benchmark of 43 planning tasks in which an operator holds hidden facts
  that the agent needs for its plan. Every hidden fact has an exact or bounded regret label, matched zero-value twins serve as controls, and
  the initial plan is scored under the true conditions, so what an agent learns is tied to the quality of its
  decision.

  \item We collect 79 operator sessions from 19 participants that show how people release information. From these sessions we build \op{}, a calibrated simulated operator, and validate its reply form, factual faithfulness and planning outcomes against the human sessions. It matches them more closely than a default LLM operator, so new agents can be evaluated repeatedly without new human dialogues.

  \item We evaluate four LLMs on all 43 tasks, comparing each model's initial plan with and without full
  hidden facts, its uptake of relevant facts against twins, its initial plan against its final plan, and two
  simulated operators. Even the best model's initial-plan loss rises from 0.03 when told every hidden fact to 0.60
  in the interactive condition, and models uncover relevant facts only 5--11 points more often than twins. Most of the loss
  comes from not asking what matters, rather than from modelling.
\end{itemize}

% --- Figure 1: one task, one relevant note, its twin --------------------------------------------------------
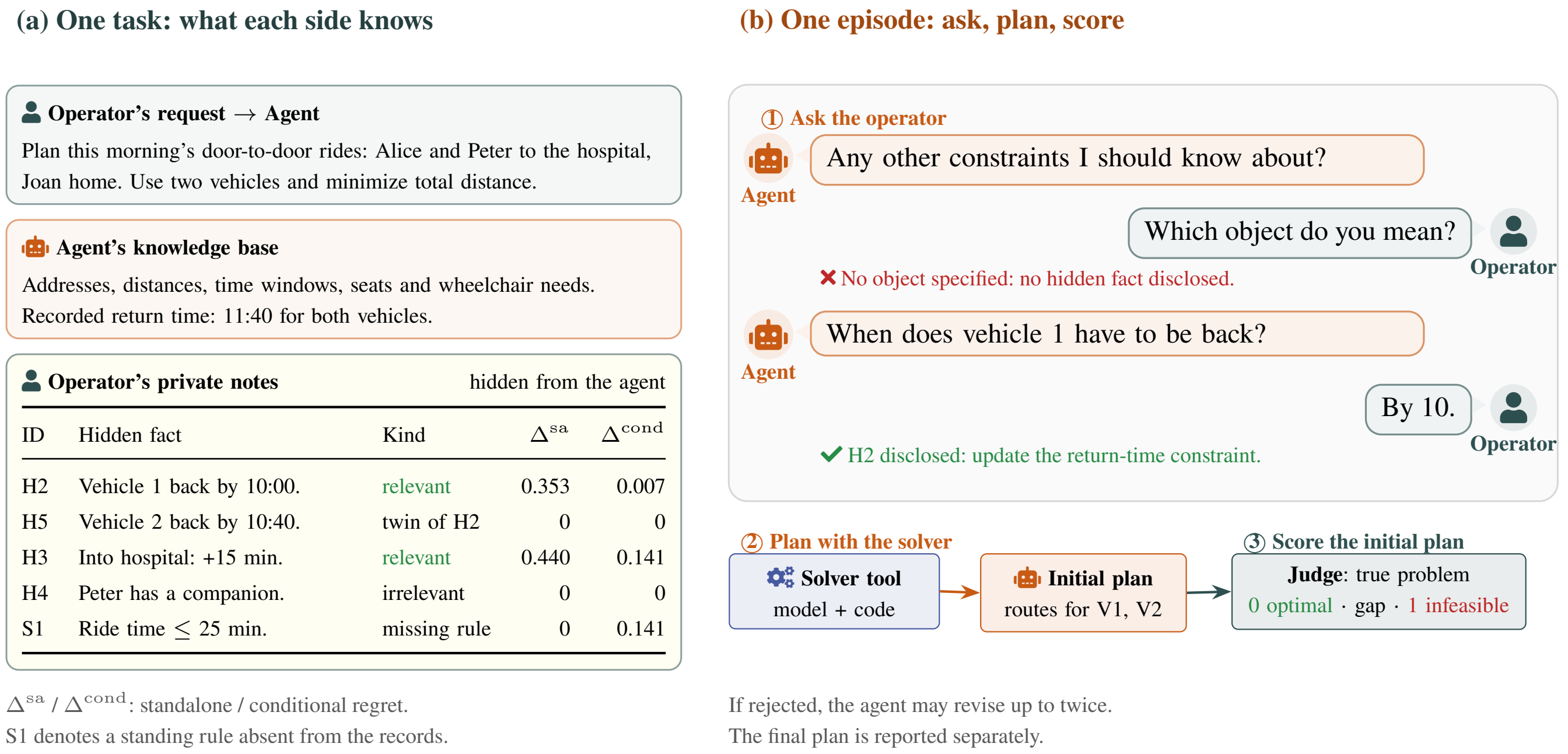
\begin{figure*}[t]
\centering
\begin{tikzpicture}[x=1cm,y=1cm,
  taskcard/.style={rounded corners=4pt, line width=0.55pt, inner sep=5pt,
    text width=7.3cm, font=\scriptsize, align=left, anchor=north west},
  chatbubble/.style={rounded corners=5pt, line width=0.6pt, inner xsep=5pt,
    inner ysep=4pt, font=\footnotesize, align=left},
  avatar/.style={circle, minimum size=0.53cm, inner sep=0pt, font=\normalsize},
  speaker/.style={font=\scriptsize\bfseries, anchor=north, inner sep=1pt}]
  % (a) Information cards. Regret values remain in aligned numeric columns.
  \node[ptitle=humancol, anchor=south west] at (0,0.05) {(a) One task: what each side knows};
  \node[taskcard, draw=humancol!55, fill=humancol!5] (brief) at (0,-0.40)
    {\textcolor{humancol}{\faUser}~\textbf{Operator's request $\rightarrow$ Agent}\\[2pt]
    Plan this morning's door-to-door rides: Alice and Peter to the hospital, Joan home. Use two vehicles and
    minimize total distance.};
  \node[taskcard, draw=agentcol!55, fill=agentcol!5] (kb) at ($(brief.south west)+(0,-0.16)$)
    {\textcolor{agentcol}{\faRobot}~\textbf{Agent's knowledge base}\\[2pt]
    Addresses, distances, time windows, seats and wheelchair needs. Recorded return time: 11:40 for both vehicles.};
  \node[taskcard, draw=humancol!55, fill=yellow!6] (notes) at ($(kb.south west)+(0,-0.16)$)
    {\textcolor{humancol}{\faUser}~\textbf{Operator's private notes} \hfill hidden from the agent\\[3pt]
    \setlength{\tabcolsep}{2pt}\renewcommand{\arraystretch}{1.15}%
    \begin{tabular*}{\linewidth}{@{\extracolsep{\fill}}l p{3.1cm} l r r@{}}
    \toprule
    ID & Hidden fact & Kind & $\Delta^{\mathrm{sa}}$ & $\Delta^{\mathrm{cond}}$ \\
    \midrule
    H2 & Vehicle~1 back by 10:00. & \textcolor{goodcol}{relevant} & 0.353 & 0.007 \\
    H5 & Vehicle~2 back by 10:40. & twin of H2 & 0 & 0 \\
    H3 & Into hospital: +15 min. & \textcolor{goodcol}{relevant} & 0.440 & 0.141 \\
    H4 & Peter has a companion. & irrelevant & 0 & 0 \\
    S1 & Ride time $\leq$ 25 min. & missing rule & 0 & 0.141 \\
    \bottomrule
    \end{tabular*}};
  % (b) A compact chat window. Every message has its own avatar and speaker label.
  \node[ptitle=agentcol, anchor=south west] at (8.20,0.05) {(b) One episode: ask, plan, score};
  \filldraw[fill=black!2, draw=black!16, rounded corners=6pt, line width=0.55pt]
    (8.20,-0.40) rectangle (17.60,-5.12);
  \node[font=\scriptsize\bfseries, text=agentcol, anchor=north west] at (8.43,-0.56)
    {\textcircled{\raisebox{-0.2pt}{1}} Ask the operator};

  \node[avatar, fill=agentcol!12, text=agentcol] (ag1) at (8.65,-1.23) {\faRobot};
  \node[speaker, text=agentcol] at (ag1.south) {Agent};
  \node[chatbubble, draw=agentcol!65, fill=agentcol!8, anchor=north west, text width=6.6cm]
    (q1) at (9.12,-0.96) {Any other constraints I should know about?};
  \fill[agentcol!8] ($(q1.north west)+(0,-0.14)$) -- ++(-0.16,-0.10) -- ++(0.16,-0.10) -- cycle;

  \node[avatar, fill=humancol!12, text=humancol] (op1) at (17.10,-2.06) {\faUser};
  \node[speaker, text=humancol] at (op1.south) {Operator};
  \node[chatbubble, draw=humancol!65, fill=humancol!7, anchor=north east]
    (a1) at (16.63,-1.79) {Which object do you mean?};
  \fill[humancol!7] ($(a1.north east)+(0,-0.14)$) -- ++(0.16,-0.10) -- ++(-0.16,-0.10) -- cycle;
  \node[font=\scriptsize, text=badcol, anchor=west] at (9.12,-2.61)
    {\faTimes~No object specified: no hidden fact disclosed.};

  \node[avatar, fill=agentcol!12, text=agentcol] (ag2) at (8.65,-3.23) {\faRobot};
  \node[speaker, text=agentcol] at (ag2.south) {Agent};
  \node[chatbubble, draw=agentcol!65, fill=agentcol!8, anchor=north west, text width=6.6cm]
    (q2) at (9.12,-2.96) {When does vehicle~1 have to be back?};
  \fill[agentcol!8] ($(q2.north west)+(0,-0.14)$) -- ++(-0.16,-0.10) -- ++(0.16,-0.10) -- cycle;

  \node[avatar, fill=humancol!12, text=humancol] (op2) at (17.10,-4.06) {\faUser};
  \node[speaker, text=humancol] at (op2.south) {Operator};
  \node[chatbubble, draw=humancol!65, fill=humancol!7, anchor=north east]
    (a2) at (16.63,-3.79) {By 10.};
  \fill[humancol!7] ($(a2.north east)+(0,-0.14)$) -- ++(0.16,-0.10) -- ++(-0.16,-0.10) -- cycle;
  \node[font=\scriptsize, text=goodcol, anchor=west] at (9.12,-4.61)
    {\faCheck~H2 disclosed: update the return-time constraint.};

  % The plan and its score remain part of the same episode.
  \node[font=\scriptsize\bfseries, text=agentcol, anchor=north west] at (8.20,-5.36)
    {\textcircled{\raisebox{-0.2pt}{2}} Plan with the solver};
  \node[font=\scriptsize\bfseries, text=humancol, anchor=north west] at (13.90,-5.36)
    {\textcircled{\raisebox{-0.2pt}{3}} Score the initial plan};
  \node[solver=2.1cm, inner sep=4pt, anchor=north west] (solve) at (8.20,-5.71)
    {\textcolor{solvecol}{\faCogs}~\textbf{Solver tool}\\model + code};
  \node[agent=2.05cm, inner sep=4pt, anchor=north west] (plan) at (11.05,-5.71)
    {\textcolor{agentcol}{\faRobot}~\textbf{Initial plan}\\routes for V1, V2};
  \node[soft=3.05cm, inner sep=4pt, anchor=north west] (judge) at (13.90,-5.71)
    {\textbf{Judge}: true problem\\
    \textcolor{goodcol}{0 optimal} $\cdot$ gap $\cdot$ \textcolor{badcol}{1 infeasible}};
  \draw[aflow] (solve.east) -- (plan.west);
  \draw[flow] (plan.east) -- (judge.west);
  % Both explanations use identical text baselines below the lowest card.
  \node[fit=(notes)(judge), inner sep=0pt] (panel-bottom) {};
  \coordinate (footerbase) at ($(panel-bottom.south west)+(0,-0.46)$);
  \node[font=\scriptsize, text=black!65, anchor=base west, inner sep=0pt] at (footerbase)
    {$\Delta^{\mathrm{sa}}$ / $\Delta^{\mathrm{cond}}$: standalone / conditional regret.};
  \node[font=\scriptsize, text=black!65, anchor=base west, inner sep=0pt]
    at ($(footerbase)+(0,-0.35)$) {S1 denotes a standing rule absent from the records.};
  \node[font=\scriptsize, text=black!65, anchor=base west, inner sep=0pt]
    at ($(footerbase)+(8.20,0)$) {If rejected, the agent may revise up to twice.};
  \node[font=\scriptsize, text=black!65, anchor=base west, inner sep=0pt]
    at ($(footerbase)+(8.20,-0.35)$) {The final plan is reported separately.};
\end{tikzpicture}
\caption{Overview of \bench{} on \texttt{hard1\_dialaride}. (a) The agent gets a brief and the records, and the operator
holds today's notes, each labelled by the solver with its regret $\Delta$. The twin H5 has the same form as H2 but
leaves the best plan unchanged. (b) The operator answers only questions about a specific object, and the agent's
initial plan is scored on the true problem. Replies are constructed for illustration.}
\label{fig:example}
\end{figure*}

% ===========================================================================
\section{\bench{}: Task Formulation}\label{sec:bench}

\bench{} measures whether an agent can ask people for the information its decision depends on.
When an LLM agent helps a human being to make a decision, some of the facts that matter are known only to that person:
they are missing from the request and from the records. A good agent does not ask about everything. It finds the
few facts that change its decision.

An \emph{episode} is one complete run of a task, from the initial request to the evaluation of the submitted
plans. It separates three roles: the \emph{agent} receives a request and incomplete records, asks questions and
submits plans; the \emph{operator} holds the missing facts and answers questions about specific objects; and the
\emph{judge} evaluates submitted plans against the true conditions.

To make the value of a question measurable, we instantiate the decision as an optimization problem. A solver can
then compute two things exactly: how much each missing fact can change the best decision, and how good the final
decision is under the true facts. Whether a question was worth asking is therefore computed by the solver, not
judged by annotators.

Section~\ref{sec:episode} specifies what the agent, operator and judge know and do, and describes how questions,
plans and revisions form an episode. Section~\ref{sec:value} defines the loss of a plan and the value of a hidden
fact, the regret that can arise when it remains unknown, and introduces twins as controls for facts that leave the
plan unchanged. Section~\ref{sec:metrics} explains how plan quality, information acquisition and recovery are
measured, and how comparison with complete information helps distinguish information acquisition from modelling.
Section~\ref{sec:construction} then describes how operational problems are turned into benchmark tasks by
specifying the initial records and hidden facts, adding controls and verifying their effects with a solver.

\subsection{Setting}\label{sec:episode}

The three roles have different responsibilities and access to information.
\begin{itemize}
  \item The \emph{agent} is the LLM under test, run in our planning harness. It receives a short request from the operator, the \emph{brief},
  and a \emph{knowledge base}: the task's existing records, such as addresses, distances, time windows and vehicle
  capacities. It can ask the operator questions, write a model and solve it, and submit plans.
  \item The \emph{operator} is a human participant or a simulated operator. It holds a few \emph{notes}: facts about
  today that the records do not show. Each note is one sentence about one object, such as ``vehicle~1 must be back
  at the depot by 10:00''. A note may change a number in the records, take an object out of use, add a rule the
  records leave out, or change what the plan should optimize.
  \item The \emph{judge} holds the true problem, that is, the knowledge base with all notes applied. It checks
  every submitted plan against this problem. The agent never sees the judge's result.
\end{itemize}

An episode runs as follows. The operator sends the brief, which states the goal (``plan this morning's rides,
shortest total distance'') but none of the notes. The agent then asks questions and runs its solver. The operator
may send at most 15 messages. The agent is told that its initial plan will be carried out and scored, so it should
ask before it plans. We call this plan the \emph{initial} plan. If the operator rejects it, the agent may revise it
up to twice. We call the last plan the \emph{final} plan and score it separately.

The operator answers only questions about a specific object. Otherwise the agent could ask ``any other
constraints?'' and receive every note at once. Human participants are
asked to follow this rule. The simulated operator enforces it and asks the agent to be more specific when a
question covers more than three objects.

Figure~\ref{fig:example} shows an example: panel (a) gives what each side knows, panel (b) one episode. The agent
plans door-to-door rides with two vehicles. The operator
holds five notes (S1, H2, H3, H4, H5). Only H2, H3 and S1 change the best plan. H5 looks just like H2,
but the same plan is best whether it holds or not. Under this rule, ``any other constraints?'' reveals nothing,
while ``when does vehicle~1 have to be back?'' reveals H2 and changes the plan, as in the episode of Figure~\ref{fig:example}b. The rest of this section defines
``changes the plan'' precisely.

\subsection{Plan Loss and the Value of a Fact}\label{sec:value}

The tested agent sees a request, the records and the operator's answers, and it
 builds its own model from them. Behind each task, however, we keep a reference model that only the judge
uses, and this model is what makes the value of a fact computable. A \emph{plan} is a decision written in the
format the task fixes, for example a route and a timetable for each vehicle. Let $\mathcal{X}$ be the set of all
such plans. A problem $\theta$ specifies which plans are feasible, $X(\theta)\subseteq\mathcal{X}$, and the cost
$f(x;\theta)$ of each plan. Its optimal value is
\begin{equation}
  z(\theta)=\min_{x\in X(\theta)} f(x;\theta).
  \label{eq:v2-opt}
\end{equation}
We write everything for minimization. For maximization, the differences below change sign. Every problem we
solve is feasible and has a nonzero optimal value.

Let $H$ be the set of hidden items of a task. For $S\subseteq H$, $\theta_S$ is the problem obtained by starting
from the agent's knowledge base and applying the true versions of the hidden items in $S$ only: every other hidden
item keeps the value the knowledge base gives it, and a rule the knowledge base lacks stays absent. Thus
$\theta_\emptyset$ is the problem exactly as the knowledge base states it, before any note is applied, and
$\theta_H$ is today's true problem, with every note applied. In Figure~\ref{fig:example}, vehicle~1 must be back by
11:40 in $\theta_\emptyset$ and by 10:00 in $\theta_H$. A hidden item can change a number in the model (48\% of
all hidden items), take an object out of use (18\%), add a constraint (25\%) or change the objective (9\%).
$\theta_S$ is the problem an agent would face if it had learned exactly the hidden items in $S$.

To score a plan $x$ under a problem $\theta$, the judge puts the plan itself into $\theta$:
\begin{equation}
  \ell(x;\theta)=\begin{cases}
  1 & x\notin X(\theta),\\
  \min\Bigl\{1,\ \dfrac{f(x;\theta)-z(\theta)}{|z(\theta)|}\Bigr\} & x\in X(\theta).
  \end{cases}
  \label{eq:v2-loss}
\end{equation}
An infeasible plan, or one not in the task's format, lies outside $X(\theta)$ and receives loss 1. Plans
submitted in an episode are scored under the true problem, $\ell(x;\theta_H)$. A plan that
cannot be carried out loses everything; otherwise its loss is the fraction by which it falls short of the best
plan, capped at 1. The judge computes $f$ from the decision of the plan proposed by the agent.

How much does it cost not to know a hidden item? An agent that has not asked about $h$ does not know which of two
problems it faces, the one without $h$ or the one with $h$, but it must still commit to one plan. Whatever $h$
does, whether it changes a number, takes an object out of use, adds a constraint or changes the objective, it
turns one problem into the other. In Figure~\ref{fig:example}, H2 moves vehicle~1's return from 11:40 to 10:00,
and S1 adds a 25-minute ride limit that the knowledge base lacks. A plan built for the problem without $h$ may
fail or do badly in the problem with $h$; a plan built for the problem with $h$ may be worse than necessary if $h$
does not hold. The cost of not knowing $h$ is the loss of the safest plan, the one whose worse case is best.
For a hidden item $h$ and a set
$S\subseteq H\setminus\{h\}$ of other hidden items whose true versions are known, we define the \emph{regret}
\begin{equation}
  \Delta_h(S)=\min_{x\in\mathcal{X}}\ \max\bigl\{\ell(x;\theta_S),\ \ell(x;\theta_{S\cup\{h\}})\bigr\},
  \label{eq:v2-delta}
\end{equation}
known as minimax regret~\citep{savage1951theory}.  $\Delta_h(S)=0$ when one plan is optimal whether or not $h$ holds, so asking about $h$
cannot improve the decision. A large $\Delta_h(S)$ means that every plan made without $h$ can go badly wrong.

% \textcolor{blue}{\textbf{Comment.} This definition takes the maximum over two scenarios, but
% Appendix~\ref{app:delta} says some earlier labels use more than two values for $h$. If those labels use all
% recorded values, define the maximum over that scenario set here and present the two-scenario formula as a special case.}

The regret depends on what else the agent already knows. We report it at the two extremes,
\begin{equation}
  \Delta^{\mathrm{sa}}_h=\Delta_h(\emptyset),\qquad \Delta^{\mathrm{cond}}_h=\Delta_h(H\setminus\{h\}).
  \label{eq:v2-sa-cond}
\end{equation}
The \emph{standalone} regret $\Delta^{\mathrm{sa}}_h$ measures how much $h$ matters to an agent that knows only the
records; the \emph{conditional} regret $\Delta^{\mathrm{cond}}_h$ measures how much it matters to an agent that
knows every other hidden item, so that its two problems are today's true problem $\theta_H$ and $\theta_H$ with
only $h$ reverted to the records.

Often the regret can be read off the two optimal values. Suppose $h$ leaves the objective unchanged and only
removes feasible plans, $X(\theta_{S\cup\{h\}})\subseteq X(\theta_S)$, as an added or tightened constraint or an
object taken out of use does. Then
\begin{equation}
  \Delta_h(S)=\min\Bigl\{1,\ \frac{z(\theta_{S\cup\{h\}})-z(\theta_S)}{|z(\theta_S)|}\Bigr\}.
  \label{eq:v2-gap}
\end{equation}
A plan infeasible under $\theta_{S\cup\{h\}}$ has loss 1 there. A plan feasible under it costs at least
$z(\theta_{S\cup\{h\}})$ in both problems, so its loss under $\theta_S$ is at least the ratio in
Eq.~\ref{eq:v2-gap}, and the optimal plan of $\theta_{S\cup\{h\}}$ attains this bound. When $h$ only adds feasible
plans, the same holds with the two problems swapped. In both cases the regret is the relative gap between the two
optimal values, measured against the optimum of the less constrained problem.

% \textcolor{blue}{\textbf{Comment.} The proof should give the gap capped at 1 as the lower bound on the loss.
% The loss cannot be at least the raw ratio when that ratio exceeds 1. The displayed formula already has the correct cap.}

In two situations, however, Eq.~\ref{eq:v2-gap} does not hold, and the regret must be computed from
Eq.~\ref{eq:v2-delta} directly. When $h$ moves a time window, neither feasible set contains the other. When
$h$ changes the objective, the optimal value can change although the optimal plan does not. In the candy-blending
task (\texttt{medium6\_candy\_blend}), today's optimal profit is \$2800. Without the true glucose-syrup price
it would be \$3250, and without the true maltose price \$5110. Both prices lower the optimal profit,
but the glucose-syrup price leaves the optimal blend unchanged, so its conditional regret is 0: the lower profit is
a cost no plan can avoid, not a consequence of not asking. The maltose price changes the blend, and its
conditional regret is 0.093. We therefore compute every label from Eq.~\ref{eq:v2-delta}. Where an optimum is not
proved (frontier-B), solver bounds turn the label into an interval (Appendix~\ref{app:delta}).

We call a hidden item \emph{relevant} if $\Delta^{\mathrm{sa}}_h>0$ or $\Delta^{\mathrm{cond}}_h>0$. The two regrets can
order hidden items differently. In Figure~\ref{fig:example}, H2 has $\Delta^{\mathrm{sa}}=0.353$ but
$\Delta^{\mathrm{cond}}=0.007$: once the other hidden items are known, it changes the plan only slightly. S1, the
25-minute ride limit, shows the opposite: it does not matter on its own ($\Delta^{\mathrm{sa}}=0$) but has
$\Delta^{\mathrm{cond}}=0.141$ when the other hidden items hold. A \emph{twin} is a hidden item of the same form as a
relevant one but about another object, with both regrets zero. H5 (vehicle~2 back by 10:40) is the twin of H2: it
sounds just as important, yet one plan is optimal whether or not it holds. An agent that asks about everything
collects both, while an agent that reasons about its plan asks about H2. Not every hidden item with zero regret is
a twin: H4 (Peter has a companion) has no relevant counterpart.

\subsection{Evaluation Metrics}\label{sec:metrics}

A single score for the whole episode would mix three abilities: building and solving a correct model, asking for
the facts that matter, and recovering after a rejected plan. A poor final plan could come from a failure in any of
them. We therefore score the episode at a fixed point, add a control without any missing information, and measure
asking directly.

The main score is the loss of the initial plan. Let $x_1$ be the initial plan an agent submits in an episode and
$x_{\mathrm{fin}}$ its last one. The \emph{initial-plan loss} is $\ell(x_1;\theta_H)$ (Eq.~\ref{eq:v2-loss}): 1 if the
plan cannot be carried out in today's true problem, and otherwise its optimality gap, capped at 1. Like the
quality--yield index for generated heuristics~\citep{chen2026heurigym}, it folds feasibility and quality into one
number, so we also report two parts of it separately: the \emph{infeasible rate}, the share of tasks with
$x_1\notin X(\theta_H)$, and the \emph{optimal rate}, the share with $\ell(x_1;\theta_H)=0$. We score the initial plan
because it is the only one made before any feedback. A rejection itself tells the agent that something is wrong,
so later plans reflect trial and error rather than asking. The \emph{final-plan loss} $\ell(x_{\mathrm{fin}};\theta_H)$
is reported separately as a measure of recovery. All scores are averaged over tasks.

A high initial-plan loss does not yet say whether the agent failed to ask or failed to model and solve. To separate
the two, we run every task in a second condition. In the \emph{interactive} condition described so far, the hidden
items are available only from the operator. In the \emph{all-info} condition, the agent is given every hidden
item together with the request and the knowledge base, so it starts from today's true problem. There is no operator,
and it submits a single plan $x^{\mathrm{all}}$. Its \emph{all-info loss} $\ell(x^{\mathrm{all}};\theta_H)$ measures how well
the agent models and solves when nothing is missing. The \emph{asking gap} is the initial-plan loss minus the all-info
loss, computed per task and then averaged. It is the part of the loss that comes from the hidden information. Both plans are
made before any feedback, so the comparison is fair.

Neither loss shows which facts the agent went after. For this we use the twins as controls. For a task, let $R$ be
its relevant hidden items, $T$ its twins and $D$ the hidden items the operator has disclosed before the initial plan.
The \emph{relevant recall} is $|D\cap R|/|R|$ and the \emph{twin rate} is $|D\cap T|/|T|$. Both are averaged over
the tasks that have relevant hidden items and twins. A twin reads just as important as its relevant counterpart, so
an agent that does not reason about its plan learns both at about the same rate, and a gap between the two shares
shows a preference for facts that matter. The operator also discloses notes it was not asked about, so we report a
stricter version in which $D$ contains only the hidden items that one of the agent's questions was about, as
judged by the operator. Finally, we report the number of questions the agent asks before its initial plan.

\section{Benchmark Construction}\label{sec:construction}

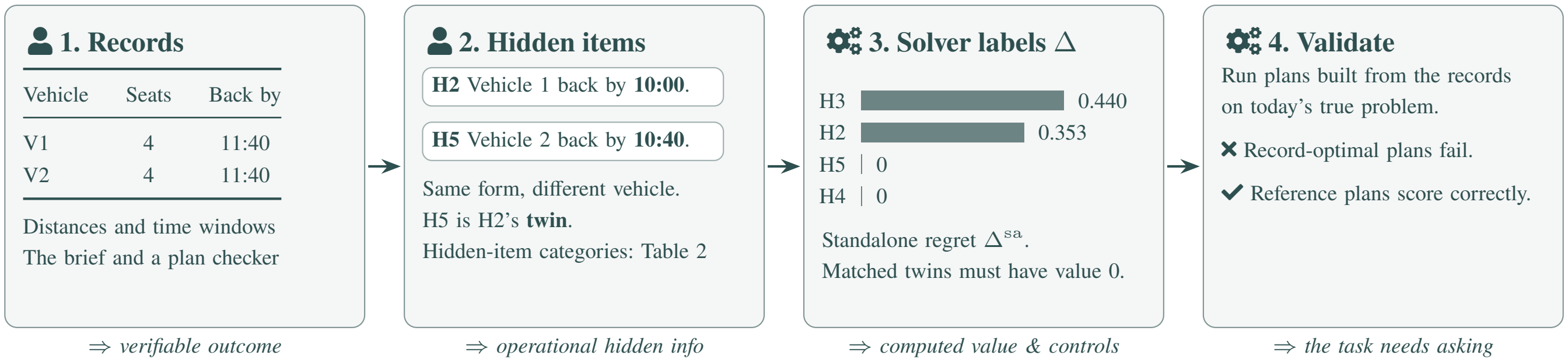
\begin{figure*}[t]
\centering
\begin{tikzpicture}[x=1cm,y=1cm, text=humancol,
  processhead/.style={font=\footnotesize\bfseries, text=humancol, anchor=north west},
  processbody/.style={font=\scriptsize, text=humancol, anchor=north west,
    inner sep=0pt, align=left, text width=3.65cm},
  processnote/.style={draw=humancol!45, fill=white, rounded corners=3pt,
    font=\scriptsize, text=humancol, inner sep=3pt, align=left,
    text width=3.2cm, anchor=north west}]
  \def\W{4.08} \def\G{0.45} \def\H{3.65}
  \foreach \i in {0,1,2,3}
    \draw[draw=humancol!60, fill=humancol!5, rounded corners=4pt]
      ({\i*(\W+\G)},0) rectangle ++(\W,-\H);
  \foreach \i in {0,1,2}
    \draw[flow] ({\i*(\W+\G)+\W+0.04},-\H/2) -- ++(\G-0.08,0);
  % ---- 1
  \node[processhead] at (0.14,-0.12) {\faUser~1.~Records};
  \node[processbody] (records) at (0.20,-0.70) {%
    \renewcommand{\arraystretch}{1.05}%
    \begin{tabular}{@{}lcc@{}}
    \toprule
    Vehicle & Seats & Back by\\ \midrule
    V1 & 4 & 11:40\\ V2 & 4 & 11:40\\ \bottomrule
    \end{tabular}};
  \node[processbody] at ($(records.south west)+(0,-0.20)$)
    {Distances and time windows\\The brief and a plan checker};
  % ---- 2
  \def\x{4.53}
  \node[processhead] at (\x+0.14,-0.12) {\faUser~2.~Hidden items};
  \node[processnote] (n1) at (\x+0.20,-0.70) {\textbf{H2} Vehicle~1 back by \textbf{10:00}.};
  \node[processnote] (n2) at ($(n1.south west)+(0,-0.17)$) {\textbf{H5} Vehicle~2 back by \textbf{10:40}.};
  \node[processbody] at ($(n2.south west)+(0,-0.22)$)
    {Same form, different vehicle.\\H5 is H2's \textbf{twin}.\\Hidden-item categories: Table~\ref{tab:hidden-kinds}};
  % ---- 3
  \def\x{9.06}
  \node[processhead] at (\x+0.14,-0.12) {\faCogs~3.~Solver labels $\Delta$};
  \foreach \name/\val [count=\i] in {H3/0.440, H2/0.353, H5/0, H4/0} {
    \node[font=\scriptsize, anchor=east] at (\x+0.6,{-0.72-\i*0.36}) {\name};
    \fill[humancol!70] (\x+0.65,{-0.72-\i*0.36-0.11}) rectangle ({\x+0.66+\val*5.2},{-0.72-\i*0.36+0.11});
    \node[font=\scriptsize, anchor=west] at ({\x+0.7+\val*5.2},{-0.72-\i*0.36}) {\val};
  }
  \node[processbody] at (\x+0.20,-2.55)
    {Standalone regret $\Delta^{\mathrm{sa}}$.\\Matched twins must have value 0.};
  % ---- 4
  \def\x{13.59}
  \node[processhead] at (\x+0.14,-0.12) {\faCogs~4.~Validate};
  \node[processbody] at (\x+0.20,-0.70)
    {Run plans built from the records\\on today's true problem.\\[4pt]
    \faTimes~Record-optimal plans fail.\\[4pt]
    \faCheck~Reference plans score correctly.};
  % criteria
  \foreach \i/\txt in {0/verifiable outcome, 1/operational hidden info, 2/{computed value \& controls}, 3/the task needs asking}
    \node[font=\scriptsize\itshape, text=humancol, anchor=north, align=center,
      text width=\W cm, inner sep=0pt, outer sep=0pt]
      at ({\i*(\W+\G)+\W/2},{-\H-0.12}) {$\Rightarrow$ \txt};
\end{tikzpicture}
\caption{How a task is built (dial-a-ride example). Steps 1--2 are written by the task author, or by a script for the
frontier splits. Steps 3--4 are run by the solver. Below each step: the design criterion it secures.}
\label{fig:pipeline}
\end{figure*}

We build each task from an optimization instance and a set of hidden operational facts. The construction
specifies the request and records available to the agent, the additional facts held by the operator, and the
complete problem used by the judge to score plans. As shown from left to right in Figure~\ref{fig:pipeline}, the
four steps prepare the records and checker, introduce hidden items and controls, compute their value labels, and
validate their effects on the plan. Together, these steps provide operational hidden information, solver-computed
values, matched controls and a verifiable outcome.

% --- Figure 2: how a task is built --------------------------------------------------------------------------

First, a task starts from an optimization instance: a textbook problem, Solomon's r101 vehicle-routing
instance~\citep{solomon1987algorithms}, or an instance from FrontierOR, ORAgentBench, OR-Library or
MIPLIB~\citep{frontieror,oragentbench,beasley1990orlibrary,gleixner2021miplib}. Its data are written as the knowledge
base, the operator's brief states the goal, and a task-specific checker scores any plan under the true specification.
This step secures a \emph{verifiable outcome}.

Second, the author writes the hidden items. They are not drawn at random. Each one is a fact of a kind that working operators hold and records miss.
Table~\ref{tab:hidden-kinds} sorts the hidden information into five kinds: basic information that was never written down, sudden
accidents, new information that arrived after the records were made, system updates not yet reflected in the
records, and human errors in the records. The kind says why the records miss a hidden item. It is independent of the
item's effect on the model, defined in Section~\ref{sec:value}, and the table tags each example with that effect. Hidden items are written for specific objects of the instance, so that the
agent can reach each one with a specific question. Some hidden items get twins of the same form about other
objects. The frontier tasks generate both from shared templates. This step secures \emph{operational hidden
information}: each hidden item is a fact about today or an unwritten rule, not a missing piece of a formulation template.

\begin{table}[t]
\centering
\small
\setlength{\tabcolsep}{3pt}
\begin{tabular}{@{}l@{\hspace{7pt}}p{0.66\columnwidth}@{}}
\toprule
Category & Example from the task suite \\
\midrule
Basic information & A standing rule staff know but the records omit: ``our vans hold 60 crates at most'' (S\#;
constraint). \\
Sudden accident & ``One truck broke down this morning, so we can run 3 truck trips at most today'' (value); ``the road from
warehouse~1 to warehouse~4 is closed today'' (availability). \\
New information & ``The hotel chain ordered 600\,kg of Value candy'' (constraint); ``CA1502 picked up a tailwind,
now expected at 08:15'' (objective). \\
System update & ``The grid company has new ramping rules: at most 20\,MW an hour'' (constraint); a price change
this week (objective). \\
Human error & A record nobody updated: ``C07's certificate for routine nursing checks has lapsed'' while the records
still list C07 for them (availability). \\
\bottomrule
\end{tabular}
\caption{The five categories of hidden information. Parentheses give each example's effect on the model.}
\label{tab:hidden-kinds}
\end{table}

Third, for every hidden item, the solver computes the standalone and conditional labels of Eq.~\eqref{eq:v2-delta}. A hidden item's kind
says how it could arise in practice, while its label says whether it changes the plan. Because the labels are
computed, this step secures \emph{computed value}, and checking that twins come out at zero secures \emph{matched
controls}.

Fourth, the construction checks whether plans that are optimal under incomplete knowledge remain optimal
today. A solver bound can exclude all such plans, whereas checking one stored solution shows only that this
solution fails. If an optimal plan survived without the relevant hidden items, the agent could do well without asking.

Table~\ref{tab:tasks} summarizes the 43 tasks, which come in five splits of growing size. Its item counts include
the missing rules (S\#), and Appendix~\ref{app:cards} lists every task.
\begin{itemize}
  \item \emph{Easy, medium and hard} (24 tasks) are textbook-size problems that a solver handles in seconds.
  They differ in how much modelling the agent must do and how the hidden items interact.
  In \emph{easy} tasks the knowledge base lists six or seven candidate plans, so the agent only has to pick one.
  Two hidden items change the plan.
  In \emph{medium} tasks the agent writes a small LP or MILP itself (transportation, assignment, covering, cutting).
  Two or three hidden items change the plan, each with a twin.
  \emph{Hard} tasks are routing and scheduling problems (dial-a-ride, job shop, unit commitment, crew duties), in
  which the value of one hidden item depends on what else is known and a missed item often makes the plan
  infeasible.
  \item The \emph{frontier} splits (19 tasks) are larger problems with many named objects (flights, vessels, trips,
  work orders, clients), built by a scripted pipeline. In \emph{frontier-A} (17 tasks) offline solves prove
  optimality, so every $\Delta$ is exact. The agent itself works under a shorter online run limit. In
  \emph{frontier-B} (2 tasks) offline solves do not close, so $\Delta$ is a certified interval and plans are scored
  against the best known solution rather than a proved optimum.
\end{itemize}

\begin{table}[t]
\centering
\small
\setlength{\tabcolsep}{4pt}
\begin{tabular}{lrrrrr}
\toprule
Split & Tasks & Items & Rel. & Twins & Max $\Delta^{\mathrm{cond}}$ \\
\midrule
Easy & 8 & 32 & 16 & 16 & 0.235 \\
Medium & 8 & 37 & 21 & 16 & 0.667 \\
Hard & 8 & 55 & 29 & 17 & 0.484 \\
Frontier-A & 17 & 101 & 59 & 42 & 0.836 \\
Frontier-B & 2 & 10 & 6 & 4 & 0.058$^\dagger$ \\
\midrule
All & 43 & 235 & 131 & 95 & 0.836 \\
\bottomrule
\end{tabular}
\caption{The 43 tasks by split. Rel.: hidden items with $\Delta^{\mathrm{cond}}>0$. $^\dagger$Upper end of a
certified interval.}
\label{tab:tasks}
\end{table}

% ===========================================================================
\section{From Human Operators to a Calibrated Simulated Operator}\label{sec:human}

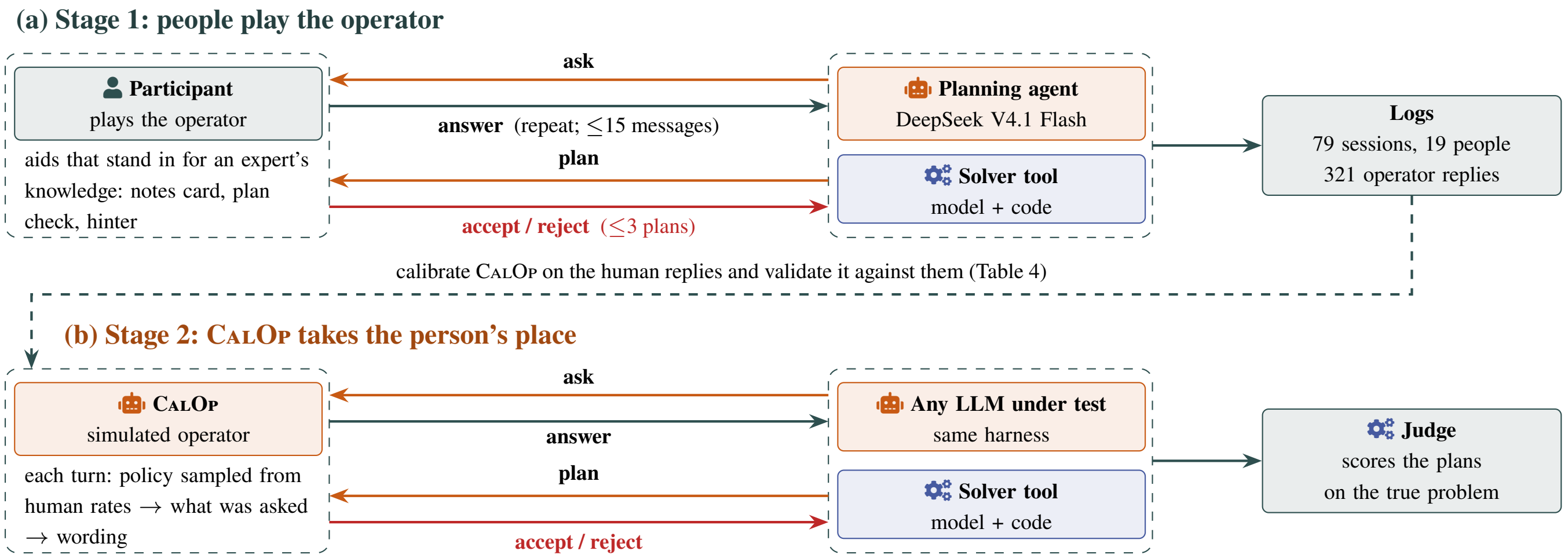
\begin{figure*}[t]
\centering
\begin{tikzpicture}[x=1cm,y=1cm]
  \def\lw{3.7} \def\rx{9.4} \def\rw{3.7}
  % ---------- row a
  \node[ptitle=humancol, anchor=south west] at (0,0.1) {(a) Stage 1: people play the operator};
  \node[soft=3.3cm, anchor=north west] (hum) at (0.1,-0.15) {\textcolor{humancol}{\faUser}~\textbf{Participant}\\plays the operator};
  \node[font=\scriptsize, anchor=north west, align=left, text width=3.5cm] (aids) at (0.1,-1.0) {aids that stand in for an
    expert's knowledge: notes card, plan check, hinter};
  \draw[grp] (0,0) rectangle (\lw,-2.1);
  \node[agent=3.3cm, anchor=north west] (ag1) at (\rx+0.1,-0.15) {\textcolor{agentcol}{\faRobot}~\textbf{Planning agent}\\DeepSeek V4.1 Flash};
  \node[solver=3.3cm, anchor=north west] (sv1) at (\rx+0.1,-1.15) {\textcolor{solvecol}{\faCogs}~\textbf{Solver tool}\\model + code};
  \draw[grp] (\rx,0) rectangle (\rx+\rw,-2.1);
  % loops a
  \draw[aflow] (\rx,-0.3) -- node[lbl, above] {\textbf{ask}} (\lw,-0.3);
  \draw[flow] (\lw,-0.6) -- node[lbl, below] {\textbf{answer}\enspace (repeat; $\le$15 messages)} (\rx,-0.6);
  \draw[aflow] (\rx,-1.45) -- node[lbl, above] {\textbf{plan}} (\lw,-1.45);
  \draw[rflow] (\lw,-1.75) -- node[lbl, below, text=badcol] {\textbf{accept / reject}\enspace ($\le$3 plans)} (\rx,-1.75);
  \node[soft=3.2cm, anchor=west] (log) at (14.35,-1.05) {\textbf{Logs}\\79 sessions, 19 people\\321 operator replies};
  \draw[flow] (\rx+\rw,-1.05) -- (log.west);
  % ---------- row b
  \def\yb{-3.6}
  \node[ptitle=agentcol, anchor=south west] at (0.55,\yb+0.1) {(b) Stage 2: \op{} takes the person's place};
  \node[agent=3.3cm, anchor=north west] (cal) at (0.1,\yb-0.15) {\textcolor{agentcol}{\faRobot}~\textbf{\op{}}\\simulated operator};
  \node[font=\scriptsize, anchor=north west, align=left, text width=3.5cm] at (0.1,\yb-1.0) {each turn: policy sampled from human
    rates $\rightarrow$ what was asked $\rightarrow$ wording};
  \draw[grp] (0,\yb) rectangle (\lw,\yb-2.1);
  \node[agent=3.3cm, anchor=north west] (ag2) at (\rx+0.1,\yb-0.15) {\textcolor{agentcol}{\faRobot}~\textbf{Any LLM under test}\\same harness};
  \node[solver=3.3cm, anchor=north west] (sv2) at (\rx+0.1,\yb-1.15) {\textcolor{solvecol}{\faCogs}~\textbf{Solver tool}\\model + code};
  \draw[grp] (\rx,\yb) rectangle (\rx+\rw,\yb-2.1);
  \draw[aflow] (\rx,\yb-0.3) -- node[lbl, above] {\textbf{ask}} (\lw,\yb-0.3);
  \draw[flow] (\lw,\yb-0.6) -- node[lbl, below] {\textbf{answer}} (\rx,\yb-0.6);
  \draw[aflow] (\rx,\yb-1.45) -- node[lbl, above] {\textbf{plan}} (\lw,\yb-1.45);
  \draw[rflow] (\lw,\yb-1.75) -- node[lbl, below, text=badcol] {\textbf{accept / reject}} (\rx,\yb-1.75);
  \node[soft=3.2cm, anchor=west] (judge) at (14.35,\yb-1.05) {\textcolor{solvecol}{\faCogs}~\textbf{Judge}\\scores the plans\\on the true problem};
  \draw[flow] (\rx+\rw,\yb-1.05) -- (judge.west);
  % calibration link
  \draw[flow, dashed] (log.south) -- (log.south |- 0,-2.75) -- node[lbl, above, pos=0.5] {calibrate \op{} on the human replies and validate it against them (Table~\ref{tab:operator})} (0.3,-2.75) -- (0.3,\yb);
\end{tikzpicture}
\caption{Study design. (a) People play the operator against a solver-equipped planning agent; the agent asks and
plans, the person answers and accepts or rejects plans, and every exchange is logged. (b) The logs calibrate
\op{}, which takes the person's place in the same loop, so that any LLM can be evaluated without new human
sessions. Orange: LLM components; teal: people and data; blue: solver.}
\label{fig:design}
\end{figure*}

Evaluating an agent needs an operator who answers it many times, over many tasks and seeds. People are too costly
for every run, and, as Table~\ref{tab:motivation-intro} shows, a default LLM operator answers unlike people. We call
this baseline \defop{}: a single prompt that gives an LLM the notes and the answering protocol. We therefore record
how people play the operator, build from these records \op{}, a human-calibrated simulated operator, and check it
against both people and \defop{}. Figure~\ref{fig:design} shows the whole design.

\paragraph{Operator study.}
In deployment, the operator is a domain expert who knows today's facts without looking them up. Our participants
cannot be experts in every task domain, so we give them what such an expert would carry in mind: hidden information notes, a check of each plan against the true conditions, and an optional hinter that points to the note a
question concerns. These aids, shown in Figure~\ref{fig:design}a, spare participants from working out the task themselves, so that they can answer as
readily as an expert would.
Every participant plays the operator of one task on a web app. They see the brief and the notes, answer the planning
agent (DeepSeek V4.1 Flash) in their own words, and accept or reject its plans. Under the answering protocol they answer only questions
about a specific object. Nineteen participants completed 79 sessions on 22
tasks, giving 321 operator replies. Appendix~\ref{app:study} describes the interface.

\paragraph{How people answer.}
Table~\ref{tab:operator} shows three regularities. (1) {People are terse}: a reply has 24.2 characters on
average, not counting spaces and punctuation, against 35.1 for \defop{}. (2) {They answer what is asked}: a reply discloses about 1.4 notes.
(3) {After a wrong plan they often say only that it is wrong}: 43\% of such replies reject or hint without
stating the correcting fact, against 7\% for \defop{}. The last point matters most for evaluation, because a
rejection that carries the fact lets an agent repair its plan without asking.

\paragraph{\op{}.}
\op{} reproduces these regularities with three steps per turn and two LLM calls. As Figure~\ref{fig:design}b shows, it takes the participant's
place in the same loop.
(1)~\emph{Policy.} Code samples the turn's behavior from the rates of the development participants: the reaction to
a plan (accept, reject, hint or correct), whether to answer in a word or two, the maximum number of facts, a length
target, and whether to number the answers.
(2)~\emph{Content.} An LLM call selects the notes that the message literally asks about. A changed fact about an
object the agent names must be stated, with numbers verbatim. A rejection targets a note flagged by the plan check.
(3)~\emph{Wording.} A second LLM call writes the reply at the target length, using real replies from other tasks as
style examples.
Both LLM calls use DeepSeek V4.1 Flash with thinking off, as does \defop{}. Unlike earlier variants
(Appendix~\ref{app:operator}), \op{} keeps no per-participant profile.

\paragraph{Validation.}
A simulated operator is useful only if it can stand in for a person, so we check three things in turn: does it
answer like a person, does it answer correctly, and does an agent reach similar plans with it as with a person?
Table~\ref{tab:operator} has one block per question and compares \op{} and \defop{} with the human column. For
the first (block a), we give \op{} the same conversation as each of the 321 human replies,
let it write its own reply, and compare its replies with the human ones on 11 measures such as length, the number
of notes disclosed and how often the fact is withheld after a wrong plan (Appendix~\ref{app:eval}). Averaged over
the measures, its gap to people is about half that of \defop{} (0.213 vs.\ 0.395), and the largest difference
disappears: after a wrong plan, people withhold the fact in 43\% of replies, \op{} in 48\% and \defop{} in only 7\%.
For the second (block b), we ask \op{} about single notes and let a scripted agent ask about every note in turn: 1.1\% of its
answers are wrong or invented, and the scripted agent reaches the optimal plan on 21 of 22 tasks. For the third (block c),
the planning agent of the study talks to each simulated operator on the same 22 tasks, three times per task. The
final plans are about as good as with people (loss 0.100 with \op{}, 0.117 with people and 0.085 with \defop{}),
and the intervals overlap. \op{} thus differs from \defop{} mainly in how it answers. RQ4 in Section~\ref{sec:rq4} shows what this changes in the evaluation of agents.

\begin{table}[t]
\centering
\small
\setlength{\tabcolsep}{3pt}
\begin{tabular}{lccc}
\toprule
 & Humans & \defop{} & \op{} \\
\midrule
\multicolumn{4}{l}{\emph{(a) Does it answer like a person? (321 replies)}} \\
Distance to humans $\downarrow$ & -- & 0.395 & \textbf{0.213} \\
Characters per reply & 24.2 & 35.1 & 25.3 \\
Notes disclosed per reply & 1.36 & 1.24 & 1.41 \\
Withholds fact after wrong plan & 0.43 & 0.07 & 0.48 \\
\midrule
\multicolumn{4}{l}{\emph{(b) Does it answer correctly?}} \\
Wrong or invented answers $\downarrow$ & -- & 0.000 & 0.011 \\
Scripted asker reaches optimum & -- & 1.00 & 0.95 \\
\midrule
\multicolumn{4}{l}{\emph{(c) Does an agent end with similar plans?}} \\
Final plan optimal & 0.815 & 0.833 & 0.773 \\
Final plan feasible & 0.915 & 0.939 & 0.924 \\
Final-plan loss & 0.117 & 0.085 & 0.100 \\
\bottomrule
\end{tabular}
\caption{Can a simulated operator stand in for people? Each block answers one question. A good operator is
close to the Humans column (79 sessions).}
\label{tab:operator}
\end{table}

% ===========================================================================
\section{Experiments}\label{sec:experiments}\label{sec:agents}

\subsection{Experimental Setting}\label{sec:experimental-setting}

\paragraph{Tasks and models.}
We evaluate four LLMs on all 43 tasks: GPT-6.1 Sol, Gemini 3.8 Flash, DeepSeek V4.1 Flash and GLM-5.3 Flash.
Each model acts as the agent in the same planning harness and calls Gurobi through a solver service. \op{} plays the operator in
the main evaluation.

\paragraph{Conditions.}
We run each model in two conditions. In the \emph{interactive} condition, the hidden facts are available only from
the operator. In the \emph{all-info} condition, the agent receives every hidden item up front and submits one plan
without an operator. This tells us how well it plans once nothing is missing. All-info is compared only with the
initial interactive plan, which is also made before any feedback, so it needs no chance to revise. To test the
operator itself, we fix DeepSeek V4.1 Flash as the model, use the 24 easy, medium and hard tasks, and compare the interactive
condition with a \emph{full-brief} condition: the operator gives every hidden item in its first message and still
reacts to plans, so both conditions allow the same revisions. Each runs with \op{} and with \defop{}, giving four
conditions and 96 episodes.

\paragraph{Protocol.}
Each episode allows at most 15 operator messages and three plans. Only the initial plan counts for the benchmark
score. Later plans measure recovery. The agent is told this in the interactive condition, and is told that it has
three attempts in the full-brief condition. Both operators follow the same answering protocol, and a check makes
sure their replies agree with the notes. The judge scores each plan on the true problem and never shows the score
to the agent. All runs use seed 0 and a solver budget of two threads and 300 seconds.

\paragraph{Metrics.}
We use the metrics of Section~\ref{sec:metrics}. For the initial plan, we report loss, the infeasible rate and the
optimal rate, together with all-info loss and the asking gap. For asking, we report relevant recall, twin rate and
the number of questions before the initial plan. For recovery, we report the same scores for the final plan.
Intervals are 95\% bootstrap intervals over tasks.

\subsection{How does access to complete information affect initial-plan quality?}\label{sec:rq1}
A model can fail a task for two reasons: it cannot model and solve the problem, or it does not learn the facts the
problem depends on. To tell the two apart, we compare each model's initial plan in the interactive condition with its initial plan
in the all-info condition. If the two are close, asking is not the bottleneck; if the first is
much worse, the loss comes from what the model did not find out.

Table~\ref{tab:main} reads from left to right. Block (a) is the benchmark score: the initial plan a model submits
in the interactive condition. Block (b) shows where the model ends after feedback, which RQ3 discusses. Block (c) gives the same
model's initial-plan loss when it is told every hidden item, and the asking gap, the part of the loss that comes
from the hidden information.
Complete information substantially improves every model's initial plan. Told every hidden item, the models reach
an initial-plan loss of 0.03--0.26; in the interactive condition, it rises to 0.60--0.98. For every model, the asking gap
(0.56--0.81) is larger than the all-info loss, so the hidden information costs more than modelling and solving.
Even the strongest model, GPT-6.1 Sol, returns an initial plan that cannot be carried out on 58\% of tasks in the
interactive condition. Most of the loss thus comes from plans that fail outright, not from plans that are merely suboptimal.

\begin{table*}[t]
\centering
\small
\begin{tabular}{lc ccc ccc cc}
\toprule
 & & \multicolumn{3}{c}{(a) Initial plan} & \multicolumn{3}{c}{(b) Final plan} &
 \multicolumn{2}{c}{(c) All-info} \\
\cmidrule(lr){3-5}\cmidrule(lr){6-8}\cmidrule(lr){9-10}
Model & \#Q & Loss\,$\downarrow$ & Infeas.\,$\downarrow$ &
Opt.\,$\uparrow$ & Loss\,$\downarrow$ & Infeas.\,$\downarrow$ & Opt.\,$\uparrow$ &
All-info loss\,$\downarrow$ & Asking gap \\
\midrule
GPT-6.1 Sol
& 8.2 & \textbf{0.603} & \textbf{58\%} & \textbf{28\%}
& \textbf{0.424} & \textbf{42\%} & 49\% & \textbf{0.029} & 0.574 \\
Gemini 3.8 Flash
& 3.9 & 0.797 & 79\% & 16\% & 0.451 & 44\% & 42\% & 0.058 & 0.739 \\
DeepSeek V4.1 Flash
& 5.9 & 0.819 & 81\% & 16\% & 0.447 & 44\% & \textbf{51\%} & 0.258 & 0.561 \\
GLM-5.3 Flash
& 0.6 & 0.979 & 98\% & 0\% & 0.623 & 58\% & 19\% & 0.173 & 0.807 \\
\bottomrule
\end{tabular}
\caption{Main results on all 43 tasks with \op{}. (a) The benchmark score: the initial plan in the interactive
condition. (b) The final plan after up to two revisions. (c) The same model's loss when told every hidden item, and the
asking gap, (a) minus (c). \#Q: questions per episode; Opt.: share of optimal plans.}
\label{tab:main}
\end{table*}

Table~\ref{tab:splits} breaks the comparison down by split and shows two sources of difficulty. On easy and
medium tasks, all-info loss is almost zero: the models can model and solve these problems, and the loss under
interaction comes from not finding out the facts. On frontier-A, some models still lose much when told every fact
(DeepSeek V4.1 Flash 0.59), so modelling and solving are also a bottleneck there. The asking gap should therefore
be read together with the all-info loss. A small gap can mean that the model asks well and does well in both
conditions, as GPT-6.1 Sol does on easy tasks (0.18 interactive, 0.00 all-info), or that it does
poorly in both, as DeepSeek V4.1 Flash does on frontier-A (1.0 and 0.59).

\begin{table}[t]
\centering
\footnotesize
\setlength{\tabcolsep}{2pt}
\begin{tabular}{lccccc}
\toprule
Model & Easy & Medium & Hard & Front.-A & Front.-B \\
\midrule
GPT-6.1 Sol & .18/.00 & .27/.00 & .67/.00 & .88/.07 & 1.0/.01 \\
Gemini 3.8 Flash & .53/.00 & .63/.00 & .88/.13 & .94/.09 & 1.0/.01 \\
DeepSeek V4.1 Flash & .40/.00 & .63/.00 & 1.0/.13 & 1.0/.59 & 1.0/.01 \\
GLM-5.3 Flash & .89/.00 & 1.0/.01 & 1.0/.02 & 1.0/.42 & 1.0/.01 \\
\midrule
\multicolumn{6}{l}{\emph{Final-plan loss, after feedback}} \\
GPT-6.1 Sol & .01 & .13 & .14 & .82 & 1.0 \\
Gemini 3.8 Flash & .01 & .01 & .51 & .77 & 1.0 \\
DeepSeek V4.1 Flash & .00 & .13 & .38 & .78 & 1.0 \\
GLM-5.3 Flash & .28 & .31 & 1.0 & .77 & .51 \\
\midrule
\multicolumn{6}{l}{\emph{Questions per episode}} \\
GPT-6.1 Sol & 3.3 & 8.3 & 4.8 & 11.2 & 15.0 \\
Gemini 3.8 Flash & 1.3 & 3.5 & 3.0 & 6.1 & 0.0 \\
DeepSeek V4.1 Flash & 2.3 & 4.8 & 6.0 & 7.7 & 8.5 \\
GLM-5.3 Flash & 0.1 & 0.0 & 0.0 & 0.7 & 6.5 \\
\bottomrule
\end{tabular}
\caption{By split. Top: initial-plan loss, interactive / all-info. Middle:
final-plan loss after feedback. Bottom: questions per episode.}
\label{tab:splits}
\end{table}

\subsection{Do models ask about facts that change the plan?}\label{sec:rq2}
A model that knows what matters should learn the relevant hidden items more often than their twins, which look
just as important but leave the plan unchanged. A model that simply asks broadly should learn both at about the
same rate. We therefore compare, before the initial plan, the share of relevant items a model learned (relevant
recall) with the same share for twins (twin rate). Table~\ref{tab:precision} counts an item in two ways.
\emph{Disclosed} counts every item the operator stated, including facts it volunteered; \emph{asked about} counts
only items that one of the model's questions targeted, so that a lucky answer does not count as good asking.

Models prefer relevant items only weakly. For the three models that ask regularly, relevant recall exceeds the
twin rate by 5--12 percentage points under either count. GLM-5.3 Flash asks almost no questions and learns almost
nothing.

\begin{table}[t]
\centering
\small
\setlength{\tabcolsep}{3pt}
\begin{tabular}{lcccc}
\toprule
 & \multicolumn{2}{c}{Disclosed} & \multicolumn{2}{c}{Asked about} \\
\cmidrule(lr){2-3}\cmidrule(lr){4-5}
Model & Relevant & Twins & Relevant & Twins \\
\midrule
GPT-6.1 Sol & 52\% & 41\% & 39\% & 27\% \\
Gemini 3.8 Flash & 39\% & 32\% & 36\% & 31\% \\
DeepSeek V4.1 Flash & 43\% & 38\% & 36\% & 29\% \\
GLM-5.3 Flash & 3\%  & 0\%  & 2\%  & 0\%  \\
\midrule
\multicolumn{5}{l}{\emph{Frontier-A only (17 tasks)}} \\
GPT-6.1 Sol & 37\% & 39\% & 24\% & 21\% \\
Gemini 3.8 Flash & 31\% & 33\% & 28\% & 32\% \\
DeepSeek V4.1 Flash & 26\% & 27\% & 15\% & 17\% \\
\bottomrule
\end{tabular}
\caption{Relevant recall and twin rate before the initial plan. Disclosed: the operator stated the item; asked
about: a question targeted it.}
\label{tab:precision}
\end{table}

On the large tasks the preference disappears. On frontier-A, all three models are told twins at least as often
as relevant items, and counting only what they asked about, the two shares differ by at most 4 points. This is not
for lack of asking: Table~\ref{tab:splits} shows that GPT-6.1 Sol asks 11.2 questions per episode on frontier-A, more than on any textbook split,
yet learns only 37\% of the relevant items. Asking more is therefore not the same as
asking the right questions. With a limited number of messages, the hard part is to work out which missing facts
could change the plan and to ask about those first.

\subsection{How much do models recover after a rejected plan?}\label{sec:rq3}
Feedback helps every model. Comparing blocks a and b of Table~\ref{tab:main}, after at most two revisions the loss
falls from 0.60--0.98 for the initial plan to 0.42--0.62 for the final one. Table~\ref{tab:splits} shows that most of the gain comes from the smaller tasks: on easy and medium tasks,
GPT-6.1 Sol, Gemini 3.8 Flash and DeepSeek V4.1 Flash end with a final loss of at most 0.13, while on frontier-A
all four models stay at 0.77--0.82. Feedback thus repairs plans when the model can solve the problem but missed
some facts, but does little when the problem itself is too hard.

Recovery, however, says little about asking. A rejection tells the agent that something about today is wrong,
and often which plan element broke, so a model can fix its plan without having asked well. GLM-5.3 Flash shows
this most clearly: it asks 0.6 questions per episode, yet its loss falls from 0.98 to 0.62 after rejections.
Likewise, GPT-6.1 Sol, Gemini 3.8 Flash and DeepSeek V4.1 Flash start far apart (0.60--0.82) but end close
together (0.42--0.45). A final-plan score therefore hides how well a model asked. We score the initial plan for
asking and report the final plan as a measure of recovery.

\subsection{How does the choice of simulated operator affect planning outcomes?}\label{sec:rq4}
Table~\ref{tab:motivation} shows that replacing \defop{} with \op{} produces mixed changes in planning outcomes.
Although \op{} better matches the observed human reply statistics in Table~\ref{tab:operator}, it yields more optimal final plans and also higher mean final loss.
This combination is possible because loss penalizes infeasibility heavily. A small number of
infeasible plans can outweigh gains in optimality elsewhere. The two metrics therefore reveal
different aspects of the outcome distribution, and neither operator consistently gives the
planning agent an advantage across the reported measures.

\begin{table*}[t]
\centering
\small
\setlength{\tabcolsep}{5pt}
\begin{tabular}{llrccccc}
\toprule
 & & & \multicolumn{2}{c}{Initial plan} & \multicolumn{3}{c}{Final plan} \\
\cmidrule(lr){4-5}\cmidrule(lr){6-8}
Condition & Operator & \#Q & Infeasible & Loss & Infeasible & Optimal & Loss \\
\midrule
Interactive & \defop{} & 4.67 & 14/24 & 0.594 & 3/24 & 17/24 & 0.136 \\
Interactive & \op{} & 4.33 & 16/24 & 0.676 & 4/24 & 20/24 & 0.167 \\
\midrule
Full brief + feedback & \defop{} & 0.08 & 2/24 & 0.083 & 2/24 & 22/24 & 0.083 \\
Full brief + feedback & \op{} & 0.04 & 1/24 & 0.042 & 1/24 & 23/24 & 0.042 \\
\bottomrule
\end{tabular}
\caption{Interactive and full-brief conditions with each operator (DeepSeek V4.1 Flash, 24 tasks).}
\label{tab:motivation}
\end{table*}

The much larger contrast is between asking for information and receiving the full brief.
Under either operator, providing the notes up front makes feasible initial plans common, whereas
interaction leaves many initial plans infeasible. The information gap thus persists across the two
simulators. Both share code-enforced answering rules, which may constrain their behavioral differences,
although isolating that effect would require an ablation. The differing instructions about revisions
also limit a causal interpretation of the information comparison.

These results support evaluating operator behavior and planning outcomes together. Reply statistics
describe how information is released, while the plan metrics measure how an agent
performs within that interaction. In this one-model, one-seed control, a matched human condition
would be needed to determine how either simulator changes scores relative to human interaction.

% ===========================================================================
\section{Related Work}\label{sec:related}

\begin{table*}[t]
\centering
\small
\setlength{\tabcolsep}{6pt}
\begin{tabular}{lccccc}
\toprule
Benchmark & Hidden & Human & Plan & Value of & Zero-value \\
 & info & data & scored & an item & controls \\
\midrule
NL4Opt, OptMATH, OPT-Engine & $\times$ & $\times$ & \checkmark & -- & $\times$ \\
HeuriGym & $\times$ & $\times$ & \checkmark & -- & $\times$ \\
QuestBench & \checkmark & $\times$ & $\times$ & required var. & $\times$ \\
OR-Clarify & \checkmark & $\times$ & $\times$ & annotated & $\times$ \\
$\tau$-bench & \checkmark & $\times$ & \checkmark & -- & $\times$ \\
Collab.\ Battleship & \checkmark & \checkmark & \checkmark & EIG (state) & $\times$ \\
\bench{} (ours) & \checkmark & \checkmark & \checkmark & regret (plan) & \checkmark \\
\bottomrule
\end{tabular}
\caption{Comparison with related benchmarks.}
\label{tab:related}
\end{table*}

Table~\ref{tab:related} places \bench{} among related benchmarks along five properties: whether information is withheld
from the agent, whether people played the answering role, whether the agent's plan or solution is scored,
how the value of a missing item is defined, and whether zero-value controls are included.

\paragraph{Clarification and information seeking.}
QuestBench~\citep{li2025questbench} casts missing information as an underspecified constraint satisfaction problem
with one missing variable. Active task disambiguation~\citep{kobalczyk2025active} and EVPI-based question
ranking~\citep{rao2018learning} choose questions by expected information value. Ask-before-Plan~\citep{zhang2024askbeforeplan}
studies clarification in travel planning. Closest to us, OR-Clarify~\citep{orclarify2026} hides formulation slots
of optimization problems, grades their severity by annotation, and checks with an LLM judge whether they were
recovered. We hide numerical and rule information as well, assign regret labels that are exact or bounded, include
$\Delta=0$ controls, and score submitted plans under the true task conditions. \citet{drossman2026conversation} study
conversational refinement of a solution with LLM stakeholders.
Collaborative Battleship~\citep{grand2026shoot} is closest in method: it collects human games, compares human and LM
askers, and scores questions by expected information gain (EIG) about the hidden board. EIG values information about
the state; our $\Delta$ values information about the decision, and the twins are hidden items with information about the
state but none about the decision. Its finding that high-EIG questions alone do not raise game performance is in
line with our separation of asking and planning.

\paragraph{LLMs for optimization modelling.}
NL4Opt, OptiMUS, ORLM, LLMOPT and OptMATH~\citep{ramamonjison2023nl4opt,ahmaditeshnizi2024optimus,huang2025orlm,jiang2025llmopt,lu2025optmath}
assume a complete problem statement. Our tasks reuse such instances (e.g.\ IndustryOR) and remove information.
OPT-Engine~\citep{chen2026optengine} scales instance size under full information and finds that added non-canonical
constraints, rather than wording or objective changes, cause most failures. Our hidden notes are such constraints,
held back. HeuriGym~\citep{chen2026heurigym} scores LLM-written heuristics by feasibility and quality against expert
solutions in a feedback loop. We score plans in the same way but withhold part of the problem.

\paragraph{Simulated users.}
$\tau$-bench~\citep{yao2024taubench} and its successors evaluate agents against LLM-simulated users. Comparisons with
real users find simulators over-cooperative and stylistically uniform, inflating agent
success~\citep{zhou2026sim2real,seshadri2026lost,dou2025simulatorarena,wang2025humanvsagent}. We calibrate the
operator on human sessions from the same tasks and validate it on the benchmark outcome.

% ===========================================================================
\section{Discussion and Conclusion}\label{sec:discussion}

\paragraph{Implications for agents.}
The results point to three open problems.
(1)~\emph{Deciding what to ask.} Told every hidden item, the models reach an initial-plan loss of 0.03--0.26; in the
interactive condition, it rises to 0.60--0.98, and they learn relevant items only 5--12 points more often than twins.
Asking more does not fix this: on frontier-A, GPT-6.1 Sol asks 11.2 questions per episode and is still told twins
as often as relevant items. The solver that scores a plan could also rank candidate questions, as expected
information gain ranks questions in information-seeking games~\citep{grand2026shoot}. A regret-guided asker is a
natural baseline.
(2)~\emph{Solving at scale.} On frontier-A, some models lose much even when told everything (DeepSeek V4.1 Flash
0.59), so modelling large instances remains a second bottleneck.
(3)~\emph{Evaluating the initial plan.} A rejection tells the agent that something is wrong, so final plans move
close together (0.42--0.45 for three models), even for a model that hardly asks. Benchmarks of interactive
planning should score the plan made before any feedback and report the final plan separately, as recovery.

% \paragraph{Limitations and future work.}
% First, the human study is small: 19 participants on 22 tasks, assisted by notes, hints and a plan check, so their
% replies need not match those of unaided domain experts, and no person has played the frontier tasks. Sessions with
% practitioners and on the large tasks would show how far \op{} carries over. Second, \op{} is calibrated and
% validated on the same pooled replies, and the operator study used a different solver setup (CBC) from the
% benchmark (Gurobi). Third, the four-model evaluation uses one seed and \op{} as the only operator, and the operator
% comparison covers one model on the 24 smaller tasks; more seeds and models would narrow the intervals. Fourth,
% relevant recall and twin rate rely on the operator's record of which items it disclosed and which item a question
% targeted. Finally, the tasks fix a message budget and require questions about specific objects, and frontier-B
% plans are scored against the best known solution rather than a proved optimum. Open-ended dialogue and decisions
% beyond optimization are natural extensions.

\paragraph{Conclusion.}
We introduced \bench{}, a benchmark of 43 planning tasks in which an operator holds hidden facts the agent needs,
a solver labels the value of every fact, and the plan is scored under the true conditions. From 79 human sessions
we built \op{}, a simulated operator that answers like people, including withholding the correcting fact after a
wrong plan, which a default LLM operator almost never does. Evaluating four LLMs, we find that current models can
solve a fully stated problem but rarely find out what the problem is: their initial plan cannot be carried out on
58--98\% of tasks in the interactive condition, and they barely prefer the facts that matter over look-alikes that do not.
Feedback hides much of this gap, so the initial plan is what measures asking. We hope \bench{} helps to build agents
that ask what matters.

% ===========================================================================
\bibliography{refs}

@article{orclarify2026,
  title   = {Ask Before You Optimize: Dynamic Pre-Formulation Clarification for Interactive Optimization},
  author  = {Ge, Sihan and Lin, Yichen and Zhou, Chenyu and Lin, Jianghao and Yao, Tao and Ge, Dongdong},
  journal = {arXiv preprint arXiv:2609.05258},
  year    = {2026}
}

@inproceedings{li2025questbench,
  title     = {{QuestBench}: Can {LLMs} Ask the Right Question to Acquire Information in Reasoning Tasks?},
  author    = {Li, Belinda Z. and Kim, Been and Wang, Zi},
  booktitle = {Advances in Neural Information Processing Systems, Datasets and Benchmarks Track},
  year      = {2025}
}

@inproceedings{kobalczyk2025active,
  title     = {Active Task Disambiguation with {LLMs}},
  author    = {Kobalczyk, Katarzyna and Astorga, Nicolas and Liu, Tennison and van der Schaar, Mihaela},
  booktitle = {International Conference on Learning Representations},
  year      = {2025}
}

@inproceedings{zhang2024askbeforeplan,
  title     = {Ask-before-Plan: Proactive Language Agents for Real-World Planning},
  author    = {Zhang, Xuan and Deng, Yang and Ren, Zifeng and Ng, See-Kiong and Chua, Tat-Seng},
  booktitle = {Findings of the Association for Computational Linguistics: EMNLP 2024},
  pages     = {10836--10863},
  year      = {2024}
}

@inproceedings{rao2018learning,
  title     = {Learning to Ask Good Questions: Ranking Clarification Questions using Neural Expected Value of Perfect Information},
  author    = {Rao, Sudha and Daum{\'e} III, Hal},
  booktitle = {Proceedings of the 56th Annual Meeting of the Association for Computational Linguistics (Volume 1: Long Papers)},
  pages     = {2737--2746},
  year      = {2018}
}

@article{drossman2026conversation,
  title   = {Let's Have a Conversation: Designing and Evaluating {LLM} Agents for Interactive Optimization},
  author  = {Drossman, Joshua and Jacquillat, Alexandre and Martin, S{\'e}bastien},
  journal = {arXiv preprint arXiv:2604.02666},
  year    = {2026}
}

@inproceedings{ramamonjison2023nl4opt,
  title     = {{NL4Opt} Competition: Formulating Optimization Problems Based on Their Natural Language Descriptions},
  author    = {Ramamonjison, Rindranirina and Yu, Timothy and Li, Raymond and Li, Haley and Carenini, Giuseppe and
               Ghaddar, Bissan and He, Shiqi and Mostajabdaveh, Mahdi and Banitalebi-Dehkordi, Amin and Zhou, Zirui and
               Zhang, Yong},
  booktitle = {Proceedings of the NeurIPS 2022 Competitions Track},
  series    = {Proceedings of Machine Learning Research},
  volume    = {220},
  pages     = {189--203},
  year      = {2023}
}

@inproceedings{ahmaditeshnizi2024optimus,
  title     = {{OptiMUS}: Scalable Optimization Modeling with ({MI}){LP} Solvers and Large Language Models},
  author    = {AhmadiTeshnizi, Ali and Gao, Wenzhi and Udell, Madeleine},
  booktitle = {International Conference on Machine Learning},
  year      = {2024}
}

@article{huang2025orlm,
  title   = {{ORLM}: A Customizable Framework in Training Large Models for Automated Optimization Modeling},
  author  = {Huang, Chenyu and Tang, Zhengyang and Hu, Shixi and Jiang, Ruoqing and Zheng, Xin and Ge, Dongdong and
             Wang, Benyou and Wang, Zizhuo},
  journal = {Operations Research},
  volume  = {73},
  number  = {6},
  pages   = {2986--3009},
  year    = {2025}
}

@inproceedings{jiang2025llmopt,
  title     = {{LLMOPT}: Learning to Define and Solve General Optimization Problems from Scratch},
  author    = {Jiang, Caigao and Shu, Xiang and Qian, Hong and Lu, Xingyu and Zhou, Jun and Zhou, Aimin and Yu, Yang},
  booktitle = {International Conference on Learning Representations},
  year      = {2025}
}

@inproceedings{lu2025optmath,
  title     = {{OptMATH}: A Scalable Bidirectional Data Synthesis Framework for Optimization Modeling},
  author    = {Lu, Hongliang and Xie, Zhonglin and Wu, Yaoyu and Ren, Can and Chen, Yuxuan and Wen, Zaiwen},
  booktitle = {International Conference on Machine Learning},
  year      = {2025}
}

@inproceedings{yao2024taubench,
  title     = {$\tau$-bench: A Benchmark for Tool-Agent-User Interaction in Real-World Domains},
  author    = {Yao, Shunyu and Shinn, Noah and Razavi, Pedram and Narasimhan, Karthik},
  booktitle = {International Conference on Learning Representations},
  year      = {2025}
}

@inproceedings{zhou2026sim2real,
  title     = {Mind the {Sim2Real} Gap in User Simulation for Agentic Tasks},
  author    = {Zhou, Xuhui and Sun, Weiwei and Ma, Qianou and Xie, Yiqing and Liu, Jiarui and Du, Weihua and
               Welleck, Sean and Yang, Yiming and Neubig, Graham and Wu, Sherry Tongshuang and Sap, Maarten},
  booktitle = {Conference on Language Modeling},
  year      = {2026}
}

@inproceedings{seshadri2026lost,
  title     = {Lost in Simulation: {LLM}-Simulated Users are Unreliable Proxies for Human Users in Agentic Evaluations},
  author    = {Seshadri, Preethi and Cahyawijaya, Samuel and Odumakinde, Ayomide and Singh, Sameer and Goldfarb-Tarrant, Seraphina},
  booktitle = {Proceedings of the 64th Annual Meeting of the Association for Computational Linguistics (Volume 1: Long Papers)},
  year      = {2026}
}

@inproceedings{dou2025simulatorarena,
  title     = {{SimulatorArena}: Are User Simulators Reliable Proxies for Multi-Turn Evaluation of {AI} Assistants?},
  author    = {Dou, Yao and Galley, Michel and Peng, Baolin and Kedzie, Chris and Cai, Weixin and Ritter, Alan and
               Quirk, Chris and Xu, Wei and Gao, Jianfeng},
  booktitle = {Proceedings of the 2025 Conference on Empirical Methods in Natural Language Processing},
  year      = {2025}
}

@inproceedings{wang2025humanvsagent,
  title     = {Human vs. Agent in Task-Oriented Conversations},
  author    = {Wang, Zhefan and Geng, Ning and Guo, Zhiqiang and Ma, Weizhi and Zhang, Min},
  booktitle = {Proceedings of the 2025 Annual International ACM SIGIR Conference on Research and Development in
               Information Retrieval in the Asia Pacific Region (SIGIR-AP)},
  pages     = {133--142},
  year      = {2025}
}

@article{savage1951theory,
  title   = {The Theory of Statistical Decision},
  author  = {Savage, Leonard J.},
  journal = {Journal of the American Statistical Association},
  volume  = {46},
  number  = {253},
  pages   = {55--67},
  year    = {1951}
}

@article{solomon1987algorithms,
  title   = {Algorithms for the Vehicle Routing and Scheduling Problems with Time Window Constraints},
  author  = {Solomon, Marius M.},
  journal = {Operations Research},
  volume  = {35},
  number  = {2},
  pages   = {254--265},
  year    = {1987}
}

@article{frontieror,
  title   = {{FrontierOR}: Benchmarking {LLMs}' Capacity for Efficient Algorithm Design in Large-Scale Optimization},
  author  = {Kong, Minwei and Jiang, Chonghe and Qu, Ao and Ouyang, Wenbin and Zeng, Zhaoming and Guo, Xiaotong and
             Li, Zhekai and Li, Junyi and Fan, Yi and Zheng, Xinshou and Jing, Xi and Zhang, Yikai and Liang, Zhiwei and
             Kim, Seonghoo and Yang, Runqing and Zhou, Zijian and Li, Sirui and Zheng, Han and Ying, Wangyang and
             Zheng, Ou and Wang, Chonghuan and Zhao, Jinglong and Qin, Hanzhang and Wu, Cathy and Liang, Paul Pu and
             Zhao, Jinhua and Wang, Hai},
  journal = {arXiv preprint arXiv:2605.25246},
  year    = {2026}
}

@article{oragentbench,
  title   = {{ORAgentBench}: Can {LLM} Agents Solve Challenging Operations Research Tasks End to End?},
  author  = {Li, Jiajun and Cai, Mingshu and Li, Yixuan and Ding, Yu and Hou, Ran and Nie, Guanyu and Han, Xiongwei and
             Wang, Wanyuan},
  journal = {arXiv preprint arXiv:2606.19787},
  year    = {2026}
}

@article{beasley1990orlibrary,
  title   = {{OR-Library}: Distributing Test Problems by Electronic Mail},
  author  = {Beasley, John E.},
  journal = {Journal of the Operational Research Society},
  volume  = {41},
  number  = {11},
  pages   = {1069--1072},
  year    = {1990}
}

@article{gleixner2021miplib,
  title   = {{MIPLIB} 2017: Data-Driven Compilation of the 6th Mixed-Integer Programming Library},
  author  = {Gleixner, Ambros and Hendel, Gregor and Gamrath, Gerald and Achterberg, Tobias and Bastubbe, Michael and
             Berthold, Timo and Christophel, Philipp and Jarck, Kati and Koch, Thorsten and Linderoth, Jeff and
             L{\"u}bbecke, Marco and Mittelmann, Hans D. and Ozyurt, Derya and Ralphs, Ted K. and Salvagnin, Domenico and
             Shinano, Yuji},
  journal = {Mathematical Programming Computation},
  volume  = {13},
  number  = {3},
  pages   = {443--490},
  year    = {2021}
}

@inproceedings{chen2026heurigym,
  title     = {{HeuriGym}: An Agentic Benchmark for {LLM}-Crafted Heuristics in Combinatorial Optimization},
  author    = {Chen, Hongzheng and Wang, Yingheng and Cai, Yaohui and Hu, Hins and Li, Jiajie and Huang, Shirley and
               Deng, Chenhui and Liang, Rongjian and Kong, Shufeng and Ren, Haoxing and Samaranayake, Samitha and
               Gomes, Carla P. and Zhang, Zhiru},
  booktitle = {International Conference on Learning Representations},
  year      = {2026}
}

@inproceedings{grand2026shoot,
  title     = {Shoot First, Ask Questions Later? {B}uilding Rational Agents that Explore and Act Like People},
  author    = {Grand, Gabriel and Pepe, Valerio and Tenenbaum, Joshua B. and Andreas, Jacob},
  booktitle = {International Conference on Learning Representations},
  year      = {2026}
}

@inproceedings{chen2026optengine,
  title     = {{OPT-Engine}: Benchmarking the Limits of {LLMs} in Optimization Modeling via Complexity Scaling},
  author    = {Chen, Yitian and Cheng, Cheng and Sun, Yinan and Ling, Zi and Ge, Dongdong},
  booktitle = {International Conference on Machine Learning},
  year      = {2026}
}

% ===========================================================================
\appendix
\section{Task inventory and cards}\label{app:cards}
The following inventory and English task-card export were generated from repository snapshot
\texttt{ba6dbd5}. Counts use only the displayed notes, not spare answer-key items. An exact positive conditional
label is relevant; for an interval, its lower bound must be positive. The two frontier hidden items below 0.005 remain
included: \texttt{x\_warehouse}/H2 (0.0032) and \texttt{x\_watch}/S1 (0.0039).
The export preserves recorded labels and outcome descriptions. It is not a new solver certification.
In Table~\ref{tab:tasks}, Rel.\ counts positive conditional labels, using positive lower bounds for interval
labels. No 0.005 threshold is applied. Twins are the stored \texttt{irrelevant} controls. Nine other hidden items have
zero conditional value and are not in that control category.

\paragraph{Effect of each hidden item on the model.} We tagged every displayed hidden item by what it changes in the model,
using its recorded model change and the task's knowledge base: a value the records hold, the availability of an
object or option, a new constraint, or the objective. Over all 235 hidden items the counts are 112 / 43 / 58 / 21, and
one hidden item (a twin in \texttt{easy1\_dispatch}) changes nothing in the model. By split (value / availability /
constraint / objective): easy 16 / 5 / 7 / 3, medium 15 / 7 / 3 / 12, hard 31 / 5 / 13 / 6, frontier-A 46 / 26 /
29 / 0, frontier-B 4 / 0 / 6 / 0. Of the 27 S\# hidden items, 25 add a constraint and 2 tighten a value the records
already hold. Twins and the other notes have similar profiles (other notes 64 / 25 / 13 / 11, twins 46 / 18 / 20 / 10),
so a twin cannot be told apart by the kind of change it makes.

\onecolumn
\begin{longtable}{p{0.37\textwidth}lrrr}
\caption{Task inventory by split, with counts of hidden items, conditionally relevant items and controls.}
\label{tab:task-inventory}\\
\toprule
Task & Split & Items & Cond. rel. & Controls\\
\midrule\endhead
\texttt{easy1\_dispatch} & easy & 4 & 2 & 2 \\
\texttt{easy2\_laundromat} & easy & 4 & 2 & 2 \\
\texttt{easy3\_family\_trip} & easy & 4 & 2 & 2 \\
\texttt{easy4\_calf\_feed} & easy & 4 & 2 & 2 \\
\texttt{easy5\_store\_delivery} & easy & 4 & 2 & 2 \\
\texttt{easy6\_ad\_buy} & easy & 4 & 2 & 2 \\
\texttt{easy7\_audit\_staffing} & easy & 4 & 2 & 2 \\
\texttt{easy8\_device\_choice} & easy & 4 & 2 & 2 \\
\texttt{medium1\_warehouses} & medium & 4 & 2 & 2 \\
\texttt{medium2\_containers} & medium & 5 & 3 & 2 \\
\texttt{medium3\_job\_assignment} & medium & 4 & 2 & 2 \\
\texttt{medium4\_nurse\_roster} & medium & 5 & 3 & 2 \\
\texttt{medium5\_warehouse\_lease} & medium & 5 & 3 & 2 \\
\texttt{medium6\_candy\_blend} & medium & 5 & 3 & 2 \\
\texttt{medium7\_rebar\_cutting} & medium & 4 & 2 & 2 \\
\texttt{medium8\_seat\_allocation} & medium & 5 & 3 & 2 \\
\texttt{hard1\_dialaride} & hard & 6 & 3 & 2 \\
\texttt{hard2\_truck\_evan} & hard & 7 & 4 & 2 \\
\texttt{hard2b\_r101\_25} & hard & 7 & 4 & 3 \\
\texttt{hard3\_job\_shop} & hard & 7 & 4 & 2 \\
\texttt{hard4\_runway} & hard & 7 & 3 & 2 \\
\texttt{hard5\_metro\_crew} & hard & 7 & 3 & 2 \\
\texttt{hard6\_unit\_commitment} & hard & 7 & 4 & 2 \\
\texttt{hard7\_bus\_network} & hard & 7 & 4 & 2 \\
\texttt{x\_airland} & frontier-A & 8 & 4 & 4 \\
\texttt{x\_berth} & frontier-A & 5 & 3 & 2 \\
\texttt{x\_crewbus} & frontier-A & 5 & 3 & 2 \\
\texttt{x\_ebus} & frontier-A & 5 & 3 & 2 \\
\texttt{x\_field\_service} & frontier-A & 7 & 4 & 3 \\
\texttt{x\_frontier2} & frontier-A & 7 & 4 & 3 \\
\texttt{x\_gap} & frontier-A & 5 & 3 & 2 \\
\texttt{x\_gate} & frontier-A & 8 & 5 & 3 \\
\texttt{x\_homecare} & frontier-A & 8 & 5 & 3 \\
\texttt{x\_jobshop} & frontier-A & 4 & 2 & 2 \\
\texttt{x\_miplib} & frontier-A & 7 & 4 & 3 \\
\texttt{x\_parmachine} & frontier-A & 5 & 3 & 2 \\
\texttt{x\_prodroute} & frontier-A & 5 & 3 & 2 \\
\texttt{x\_unitcommit} & frontier-A & 5 & 3 & 2 \\
\texttt{x\_vrpb} & frontier-A & 7 & 4 & 3 \\
\texttt{x\_warehouse} & frontier-A & 5 & 3 & 2 \\
\texttt{x\_watch} & frontier-A & 5 & 3 & 2 \\
\texttt{x\_frontier2\_b} & frontier-B & 5 & 3 & 2 \\
\texttt{x\_miplib\_b} & frontier-B & 5 & 3 & 2 \\
\bottomrule
\end{longtable}
\twocolumn

\section{Computing regret labels}\label{app:delta}

Each label is $\Delta_h(S)$ from Eq.~\ref{eq:v2-delta}. It compares two problems, $\theta_S$ and
$\theta_{S\cup\{h\}}$, that differ only in the hidden item $h$ being tested: the standalone label uses $S=\emptyset$,
so every other hidden item keeps its knowledge-base value, and the conditional label uses $S=H\setminus\{h\}$, so
every other hidden item is at its true value. When $h$ leaves the objective unchanged and only removes or only adds
feasible plans, the label follows from the two optimal values by Eq.~\ref{eq:v2-gap}. Otherwise, when $h$ changes
the objective or moves a time window, we solve one more optimization problem, of the same type as the task's own
model: it looks for a single plan that is feasible in both problems and minimizes the larger of its two losses. If
no plan is feasible in both, the label is 1. Some earlier tasks record more than two plausible values for a hidden
item; the maximum in Eq.~\ref{eq:v2-delta} then runs over the problems for all of them.
Exact labels require proved optima. If a solve remains open, solver bounds give a regret interval. A zero label
applies to the background in which it was checked, rather than to every possible knowledge state.

The frontier tasks generate twins from shared templates. The earlier tasks also contain less strictly matched
irrelevant notes, such as H4 in Figure~\ref{fig:example}. We count only the stored {irrelevant} controls as
twins. All of them have zero standalone and zero conditional regret. We do not assume that every relevant hidden
item has its own twin.

\section{Human study and assistance}\label{app:study}
The selected data contain 79 completed protocol sessions.
They cover 22 tasks. The 321 replies are split by participant: nine participants and 161 replies in development,
ten participants and 160 replies in test. The human data do not include \texttt{hard1\_dialaride} or
\texttt{hard2b\_r101\_25}. Figure~\ref{fig:example} is therefore an illustrative task, not a transcript excerpt.

Participants see a brief and hidden notes, answer the planning agent, and accept or reject its plans. The app
disables ordinary paste actions. Each plan card can show a check under the true task conditions. The optional
Hinter can explain which question to answer and, in the checked implementation, can state a numeric answer from
the notes or reference. It must not disclose the reference optimum as an answer. The planning agent does not
receive the participant's private plan-check display directly.

Each session allows up to 15 messages and three plans. The planning agent can call a solver: it sends a model
and PuLP code, and each plan must state the result of its latest optimal run. Measured on the raw text with
whitespace removed, the median human reply has 15 characters and 40.8\% of replies contain no punctuation. The
24.2-character mean in Table~\ref{tab:operator} instead excludes spaces and punctuation.

The historical study uses sandboxed Python/PuLP with CBC and a 60-second solve limit. The main benchmark instead
uses the Gurobi service described in the experimental setup. These settings must not be treated as identical.

\section{Operator design and earlier variants}\label{app:operator}
The selected \texttt{f\_min4} operator samples reply behavior from population statistics and separates content
selection from wording.

Earlier variants added per-person profiles, calibration, style controls and additional checks. The pooled replay
distances are 0.395 for \texttt{default}, 0.350 for \texttt{pc2s}, 0.339 for \texttt{f\_v5}, 0.289 for
\texttt{f\_v6}, 0.242 for \texttt{f\_v7}, and 0.213 for \texttt{f\_min4}. These are comparisons among developed
systems, not controlled estimates of the causal effect of removing each component. They also do not establish
which version best represents a new participant population.

The planning benchmark adds an object-enumeration guard and a truth-check wrapper. The historical operator
validation does not measure the combined effect of these components. The historical probe results show that
\texttt{f\_min4} is imperfect: its oracle loop succeeds on 21 of 22 tasks. The failure involves an answer that
names a football match without stating that the person cannot attend.

In the full-brief comparison, the operator instructions still describe the notes as private even though the
agent receives them. This limits interpretation of that comparison.

\section{Evaluator definitions and uncertainty}\label{app:eval}
The 11 replay measures are character count, sentence count, colloquial rate, structured-answer rate,
information-point count, information density, extra-information ratio, hidden notes disclosed, the share of
replies that disclose at least two notes, repetition rate, and withholding after a wrong plan. Character count
excludes spaces and punctuation. Information density is 100 times the number of information points divided by
that count. The semantic annotations use a fixed model judge and rubric. They are not human ground truth.
The historical evaluator reports 5,816 annotated replies out of 5,827 across its selected sources. Metric
sample counts differ when an annotation or denominator is missing.

For metric $m$, let $a_m$ and $h_m$ be the operator and human group means. The pooled distance is
\begin{equation}
 D=\frac{1}{11}\sum_{m=1}^{11}\frac{|a_m-h_m|}{\max(|h_m|,0.05)}.
 \label{eq:operator-distance}
\end{equation}
This summary can hide distributional and per-participant differences. Paired reply errors are separate metrics.
A development/test split alone does not turn the pooled summary into held-out validation.

For closed-loop outcomes, we average sessions or seeds within a task and then average the 22 task means. Because
every task has equal weight, these rates differ slightly from the per-session rates in
Table~\ref{tab:motivation-intro}.
The table below uses 20,000 percentile bootstrap samples of those task means, with bootstrap seed 20261101.
These intervals condition on the observed participants and runs. They do not capture participant-population
uncertainty. Historical optimality means a feasible plan with relative gap at most 0.001.
\begin{table}[h]
\centering\small
\begin{tabular}{llc}
\toprule
Source & Metric & Mean [95\% task interval]\\
\midrule
Humans & Optimal & .815 [.695, .918]\\
       & Feasible & .915 [.832, .980]\\
       & Loss & .117 [.047, .201]\\
\defop{} & Optimal & .833 [.697, .955]\\
       & Feasible & .939 [.864, 1.000]\\
       & Loss & .085 [.016, .173]\\
\op{} & Optimal & .773 [.652, .894]\\
       & Feasible & .924 [.848, .985]\\
       & Loss & .100 [.035, .178]\\
\bottomrule
\end{tabular}
\caption{Task-balanced historical operator outcomes. Simulated runs use seeds 0--2; the additional default
seed 3 is excluded to keep the comparison matched. Intervals are descriptive, not equivalence tests.}
\label{tab:historical-operator-outcomes}
\end{table}

For the main evaluation in Tables~\ref{tab:main} and~\ref{tab:precision}, the task-bootstrap 95\% intervals are:
initial-plan loss GPT-6.1 Sol [0.46, 0.74], Gemini 3.8 Flash [0.67, 0.91], DeepSeek V4.1 Flash [0.70, 0.93],
GLM-5.3 Flash [0.94, 1.00]; relevant recall (disclosed) GPT-6.1 Sol [42\%, 63\%], Gemini 3.8 Flash [28\%, 50\%],
DeepSeek V4.1 Flash [34\%, 53\%]; twin rate (disclosed) GPT-6.1 Sol [30\%, 53\%], Gemini 3.8 Flash [21\%, 43\%],
DeepSeek V4.1 Flash [27\%, 50\%].

\end{document}